\documentclass{article}

\PassOptionsToPackage{numbers, sort}{natbib}
\usepackage[preprint]{neurips_2024}

\usepackage[utf8]{inputenc} 
\usepackage[T1]{fontenc}    
\usepackage{hyperref}       
\usepackage{url}            
\usepackage{booktabs}       
\usepackage{amsfonts}       
\usepackage{nicefrac}       
\usepackage{microtype}      
\usepackage{xcolor}         
\usepackage{threeparttable} 
\usepackage{tikz}
\usepackage{pgfplots}
\pgfplotsset{compat=1.16}
\usepgfplotslibrary{groupplots}

\newcommand{\todo}[1]{}
\renewcommand{\todo}[1]{{\color{red} TODO: {#1}}}

\newcommand{\maybe}[1]{}
\renewcommand{\maybe}[1]{{\color{blue} MAYBE?: {#1}}}

\usepackage{ifthen}

\newcommand{\scenabled}{0}

\ifthenelse{\equal{\scenabled}{1}}{
\newcommand{\vista}{{\scshape v}{\small i}{\scshape st}{\small a}}
\newcommand{\capvista}{ViSTa}
}
{
\newcommand{\vista}{ViSTa}
\newcommand{\capvista}{\vista}
}

\newcommand{\longvista}{{\bfseries Vi}sion-based understanding of {\bfseries S}equential {\bfseries Ta}sks}

\ifthenelse{\equal{\scenabled}{1}}{
\newcommand{\clip}{{\scshape clip}}

}{
\newcommand{\clip}{CLIP}

}

\ifthenelse{\equal{\scenabled}{1}}{
\newcommand{\viclip}{{\scshape v}{\small i}{\scshape clip}}

}{
\newcommand{\viclip}{ViCLIP}

}

\ifthenelse{\equal{\scenabled}{1}}{
\newcommand{\vlm}{\scshape vlm}
\newcommand{\vlms}{{\scshape vlm}{\small s}}

\newcommand{\capvlms}{VLMs}
}{
\newcommand{\vlm}{VLM}
\newcommand{\vlms}{VLMs}

\newcommand{\capvlms}{\vlms}
}

\ifthenelse{\equal{\scenabled}{1}}{
\newcommand{\gptfour}{\scshape gpt-4}
}{
\newcommand{\gptfour}{GPT-4}
}

\ifthenelse{\equal{\scenabled}{1}}{
\newcommand{\basalt}{\scshape basalt}
}{
\newcommand{\basalt}{BASALT}
}

\ifthenelse{\equal{\scenabled}{1}}{
\newcommand{\alfred}{\scshape alfred}
}{
\newcommand{\alfred}{ALFRED}
}

\ifthenelse{\equal{\scenabled}{1}}{

}{

}

\ifthenelse{\equal{\scenabled}{1}}{
\newcommand{\rl}{\scshape rl}

\newcommand{\rlhf}{\scshape rlhf}

}{
\newcommand{\rl}{RL}

\newcommand{\rlhf}{RLHF}

}

\ifthenelse{\equal{\scenabled}{1}}{

}{

}

\newcommand{\david}[1]{}
\renewcommand{\david}[1]{{\color{magenta} DAVID: {#1}}}

\newcommand{\mike}[1]{}
\renewcommand{\mike}[1]{{\color{orange} MIKE: {#1}}}

\usepackage{graphicx}
\usepackage{xcolor}
\usepackage{multirow}
\usepackage{subcaption}
\usepackage{amsmath} 

\title{{\capvista} Dataset: Do vision-language models\\understand sequential tasks?}

\author{Evžen Wybitul\thanks{ewybitul@ethz.ch}, Evan Ryan Gunter\thanks{See author contribution statement.}, Mikhail Seleznyov\footnotemark[\value{footnote}] \\
ML Alignment and Theory Scholars (MATS) \\
\AND David Lindner \\
Google DeepMind
}

\usepackage{cleveref}
\usepackage{listings}

\begin{document}

\maketitle

\setcounter{footnote}{0}  

\begin{abstract}
Using vision-language models ({\vlms}) as reward models in reinforcement learning holds promise for reducing costs and improving safety.
So far, {\vlm} reward models have only been used for goal-oriented tasks, where the agent must reach a particular final outcome.
We explore {\vlms}' potential to supervise tasks that cannot be scored by the final state alone. To this end, we introduce {\vista}\footnote{\url{https://github.com/Eugleo/vista-dataset}}, a dataset for evaluating \longvista.
{\vista} comprises over 4,000 videos with step-by-step descriptions in virtual home, Minecraft, and real-world environments.
Its novel hierarchical structure---basic single-step tasks composed into more and more complex sequential tasks---allows a fine-grained understanding of how well {\vlms} can judge tasks with varying complexity.
To illustrate this, we use {\vista} to evaluate state-of-the-art {\vlms}, including \clip, {\viclip}, and {\gptfour}o. We find that, while they are all good at object recognition, they fail to understand sequential tasks, with only {\gptfour}o achieving non-trivial performance.
\end{abstract}

\begin{figure}[h]
    \centering
    \includegraphics[width=\textwidth]{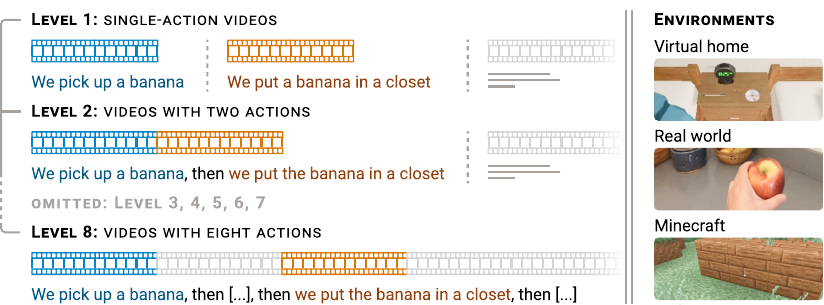}
    \caption{
        \textbf{{\capvista} is a hierarchical dataset of videos with step-by-step descriptions.}
        {\capvista} enables granular testing of task sequences in three different environments.
        Tasks are organized by number of sub-tasks into a hierarchy of 8 levels.
        Videos within levels are grouped into problem sets (\cref{fig:problem-example}) testing specific capabilities.
    }
    \label{fig:overview}
\end{figure}

\section{Introduction}

Reinforcement learning ({\rl}) can excel at tasks like games and robotics which require complex sequential decision making~\cite{alphago, tang2024deep}.
However it is difficult to hand-craft reliable reward functions for many tasks~\cite{krakovna2020specification, skalse2022defining}, and
optimizing a proxy reward function which doesn't quite capture what we want can lead to unintended---or even unsafe---behavior~\cite{halfcheetah, trickvisionclassifier, backflipandtrickyrobothand}.

How can we get the nuanced judgments {\rl} requires?
The simplest solution is to use humans.
Human supervision can indeed be used to teach tasks which cannot be precisely specified \cite{backflipandtrickyrobothand}.
However, human supervision is expensive---too expensive to teach highly complex tasks.
Also, human supervisors who cannot scrutinize every frame may be tricked by policies which appear good but actually fail to perform their task \cite{backflipandtrickyrobothand}.

Using {\vlm} supervisors instead of humans could address these problems.
{\capvlms} are much cheaper than human supervision, and show potential to be capable task supervisors.
{\capvlms} have already been used to judge whether a desired outcome has been achieved~\citep{rocamondeVisionLanguageModelsAre2023, sontakkeRoboCLIPOneDemonstration2023}.
They may be even more useful for judging whole trajectories: they could supervise sequential tasks, detect reward hacking, and provide guidance when ground truth rewards are sparse without the fragility of handcrafted reward shaping---while being cheaper and maybe even more reliable than human supervisors.
This has only just begun to be explored \cite{guanTaskSuccessNot2024}.

\begin{figure}[h]
    \centering
    \includegraphics[width=\textwidth]{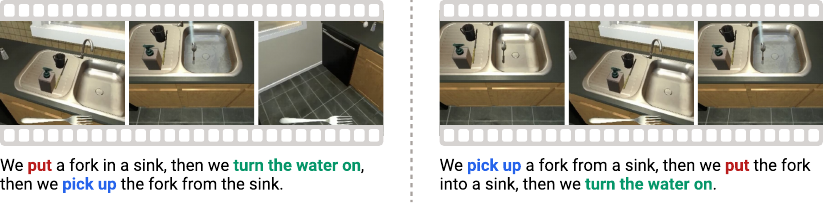}
    \caption{
        \textbf{A problem set for action-order understanding.} Problem sets are groups of videos to be matched with their descriptions. Each set targets a specific capability, e.g.\ understanding action order.
    }
    \label{fig:problem-example}
\end{figure}

To facilitate this, we present {\vista}, a hierarchical dataset for evaluating \longvista.
{\capvista} comprises over 4,000 videos of tasks with step-by-step descriptions, across 3 environments: virtual home, Minecraft, and the real world.
{\capvista} is hierarchical (\cref{fig:overview}): it has basic \emph{single-action} tasks which combine into increasingly complex \emph{multiple-action} tasks.
Video-description pairs are grouped into problem sets (\cref{fig:problem-example}) that test specific capabilities such as object recognition and understanding action order.
We use {\vista} to evaluate several {\vlms}, including the frontier model {\gptfour}o, and find them not yet capable of supervising any but the most basic tasks.

\section{Related work} \label{sec:related-work}

\paragraph{{\capvlms} as reward models.} Previous studies have extensively explored using vision~\cite{inthewildvideos, manypretrainedinclclip, trickvisionclassifier}, language~\cite{llmsupervisornovisual, lambert2024rewardbench}, and vision-language models~\cite{languageandphotodemo, clipsupervisor, rocamondeVisionLanguageModelsAre2023, sontakkeRoboCLIPOneDemonstration2023, wangRLVLMFReinforcementLearning2024, lambert2024rewardbench, guanTaskSuccessNot2024} for success detection in outcome-based tasks.
For instance, \citet{clipsupervisor} demonstrated that models like CLIP can act as reward models using goal descriptions such as “a yellow block on top of a red block.” \citet{rocamondeVisionLanguageModelsAre2023} showed that CLIP can serve as a reward model without fine-tuning for qualitative tasks like “doing the splits.” Other works~\cite{languageandphotodemo, sontakkeRoboCLIPOneDemonstration2023} leveraged VLMs to improve one-shot learning in robot policy training.
Additionally, \citet{wangRLVLMFReinforcementLearning2024} illustrated how costly frontier VLMs can generate training data for smaller reward models, akin to \rlhf.

These works focus on outcome-based supervision; even if the {\vlm} has access to the whole trajectory, in the end it only assess whether a particular outcome was reached.
However, process-based supervision --- one in which the trajectory itself undergoes detailed scrutiny --- is less fragile.
Supervisors which see only a single frame have been tricked by {\rl} policies optimized against them \cite{trickvisionclassifier}.
Additionally, \citet{guanTaskSuccessNot2024} highlighted the pitfalls of judging only outcomes and proposed using frontier VLMs to evaluate entire trajectories for undesired behaviors. 
Their results, while promising, also revealed surprising failures in seemingly capable models.
With {\vista}, we aim to make it easier to systematically assess when models succeed and fail, to determine whether they are ready to be reliable task supervisors.

\paragraph{Benchmarks and datasets for VLMs.} 
Existing benchmarks such as PCA-Bench~\cite{chenPCABenchEvaluatingMultimodal2024} incidentally test {\vlm} perception but focus on decision-making rather than evaluating task performance, and only consider non-deceptive environments. 
Important to our work is the study by \citet{wangPaxionPatchingAction2023}, showing that {\vlms} overly rely on object recognition and provided a benchmark with tasks where object recognition is insufficient. 
Building on these insights, our goal is to develop a granular, systematic evaluation of {\vlm} capabilities.

In constructing our dataset, we drew inspiration from prior benchmarks and datasets designed for {\vlms} and embodied AI systems~\cite{liBEHAVIOR1KHumanCenteredEmbodied2024, puigVirtualHomeSimulatingHousehold2018, savvaHabitatPlatformEmbodied2019, puigHabitatCoHabitatHumans2023a, shridharALFREDBenchmarkInterpreting2020}.
These datasets, aimed at training agents in virtual environments, lack the structure and granularity needed for {\vlm} evaluation, especially when our goal is to assess specific capabilities.
Nevertheless, we wanted to make use of the rich data the existing training datasets and benchmarks offer.
We based part of our dataset on videos from ALFRED~\cite{shridharALFREDBenchmarkInterpreting2020}, enhancing it by organizing videos into problem sets and often creating new videos from existing clips to test specific capabilities.

In contrast to prior work, {\vista} is focused on evaluating VLMs' ability to understand and rate performance on sequential tasks. 
Moreover, {\vista} includes data from 3 different environments, as opposed to the prior benchmarks that often have only a single environment.
In particular, we include real videos similar to those in virtual home to allow direct comparison of {\vlm} efficacy in on-distribution real-world videos vs.\ similar simulations; and we include Minecraft videos to assess {\vlm} capabilities in even more unfamiliar simulated environments.

\section{{\capvista} Dataset: hierarchy, problem sets, and data sources} \label{sec:dataset}

{\capvista} contains more than 4,000 videos and descriptions of tasks in three environments: virtual home, Minecraft, and the real world.
Many tasks in these environments can be viewed as sequences of atomic sub-tasks: ``Pick up A, go to B, place A in B, turn B on''. 
{\capvista} mirrors this structure:
\begin{enumerate}
    \item \textbf{Single-action videos (level 1):} 
    These test if a model can identify fundamental actions like ``mine a wooden block'', ``open a door'', or ``put a banana into the closet''. 
    The actions are sometimes quite complex: for example, the video ``heat up an apple'' shows the agent putting the apple in a microwave, turning it on, waiting, then picking the apple back up.
    
    \item \textbf{Multiple-action videos (levels 2 through 8):} 
    These use sequences of the fundamental actions like ``pick up an apple, then put the apple in the drawer'' to test if a model understands action order and if it notices when we swap out actions for different ones.
    The length of the sequence --- the number of actions --- depends on the level.
    Different-length tasks let us quantify how models' sequential reasoning capabilities break down.
    
\end{enumerate}

\subsection{Problem sets}

{\capvista} groups the video-description pairs into \emph{problem sets}: classification problems testing specific capabilities.
During the evaluation of a problem set, models get a video and score how well it matches each description from the problem set. We list all problem sets in \cref{app:problem-sets}. In \cref{sec:results}, we aggregate results from different problem sets by grouping them by theme:
\begin{itemize}
    \item \textbf{Objects:} Level 1 (single-action) problem sets testing object recognition.
    They contain videos such as ``We pick up an \textsc{apple}'', or ``We pick up a \textsc{hammer}''.
    \vista{} covers most objects that appear in \alfred{}.

    \item \textbf{Object properties}: Level 1 (single-action) problem sets testing whether the model can detect specific object properties ---open/closed, turned on/turned off, etc.
    The problem sets have videos such as ``We observe an \textsc{open} drawer'', or ``We observe a \textsc{closed} drawer''.

    \item \textbf{Actions}: Level 1 (single-action) problem sets in which the model needs to understand what a particular action (heating, cooling, cleaning, etc.) involves.
    The videos include ``We \textsc{heat up} a banana'', or ``We put a banana in a microwave \textsc{without turning it on}''.

    \item \textbf{General problems:} Level $n$ (multiple-action) problem sets testing general sequential task understanding, e.g. ``We open a drawer, then we pick up a banana from the drawer.''
    Models must determine which of several possible sequences of actions matches the video.

    \item \textbf{Permutation problems:} Level $n$ (multiple-action) problem sets testing whether the model can understand action order. 
    In a given problem set, the videos are permutations of the same actions, differing only in their order. 
    \Cref{fig:problem-example} illustrates such a problem set.
    
\end{itemize}

\subsection{Videos in {\vista} come from different sources}

Some videos in {\vista} are from existing datasets; most are manually filmed or edited. 

\paragraph{Virtual home.} 
\vista{} contains more than 3,000 virtual home videos of levels 1 through 8.
The videos are clips from {\alfred}~\cite{shridharALFREDBenchmarkInterpreting2020} and combinations thereof.
Different frames may come from different source videos, and so the videos have small visual discontinuities.
We've minimized these when possible, and they are likely undetectable by current {\vlms} that have low frame rates.

\paragraph{Real world.} 
We produced more than 900 videos in levels 1--5 and 8 which show the agent (us) doing tasks in the real world.
810 of those videos are reproductions of tasks from the virtual home environment.
Additionally, we created videos for tasks to which virtual home is ill-suited:
(1) Object tracking: similar objects being shuffled. 
(2) Object interactions: pinning fabric, or falsely seeming to do so. 
Finally, we also collected 200 videos of a door either opening or closing from Kinetics-700~\cite{kinetics}, to test action recognition in complex contexts.

\paragraph{Minecraft.} 
In Minecraft, \vista{} has 53 videos of levels 1 through 3. Most were created manually; the rest were sourced from the {\basalt} benchmark~\cite{milani2023bedd}. We took inspiration from \citet{yuan2023plan4mc} in choosing the set of basic actions for the single-action videos.

\section{Using ViSTa to evaluate models} \label{sec:results}

\newcommand{\fscore}{F\textsubscript{1} score}

{\capvista}'s video-description matching approach supports a variety of {\vlms}. We demonstrate it on 3 models: {\clip}~\cite{clip, chertiReproducibleScalingLaws2022, schuhmannLAION5BOpenLargescale2022}, a capable model without native support for videos; {\viclip}~\cite{viclip}, a smaller model which does support videos natively; and {\gptfour}o~\cite{hurst2024gpt}, a frontier {\vlm}.

\subsection{Experimental setup}

We use the models to score each video-description pair in a problem set, and pick the description that has the highest score as the models' predictions. The scoring process depends on the model:

\begin{itemize}
    \item \textbf{\viclip:} We use {\viclip} to produce video and description embeddings, and compute the score as their cosine similarity. 
    Note that while {\viclip} does support videos natively, it only uses 8 (equally spaced) frames from the video.

    \item \textbf{\clip:} As for {\viclip}, we compute the score as a cosine similarity between a video embedding and a description embedding.
    However, {\clip} only supports static images, so we embed videos by averaging the separate embeddings of 32 equally spaced frames, completely neglecting the order of the frames. 
    
    \item \textbf{{\gptfour}o:} We give {\gptfour}o 16 frames from the video and prompt it to describe them, write a summary, and then score how well the summary matches each of the descriptions in the problem set. See \cref{app:experimental-setup-details} for a full list of prompts.
\end{itemize}

Instead of using the raw model scores, we first standardize them, as this greatly improves model performance.
Concretely, for a problem with classes $1,\ldots,K$, videos $1,\ldots,V$, and intermediate model scores $s^\text{pre}_{v, k}$, we compute the final scores as
\begin{align*}
    s^\text{final}_{v, k} = \frac{s^\text{norm}_{v, k} - \mu}{\sigma},
    \qquad s^\text{norm}_{v, k} = \frac{\exp s^{\text{pre}}_{v, k}}{\sum_{i \in K} \exp s^{\text{pre}}_{v, i}}.
\end{align*}
Here, $\mu$ and $\sigma$ are the mean and standard deviation computed from normalized scores for the $k$-th class of videos.\footnote{Results did not meaningfully differ if we excluded the video being tested from the computation of $\mu$ and $\sigma$.}

\subsection{{\capvlms} are not ready to supervise sequential tasks}

Overall, we see results that confirm previous findings about \vlms' capabilities. They suggest that the models cannot be used as task supervisors yet. We have additional figures in \cref{app:full-results}.

\begin{figure}[ht]
    \centering
    \begin{subfigure}[t]{0.49\textwidth}
        \includegraphics[width=\textwidth]{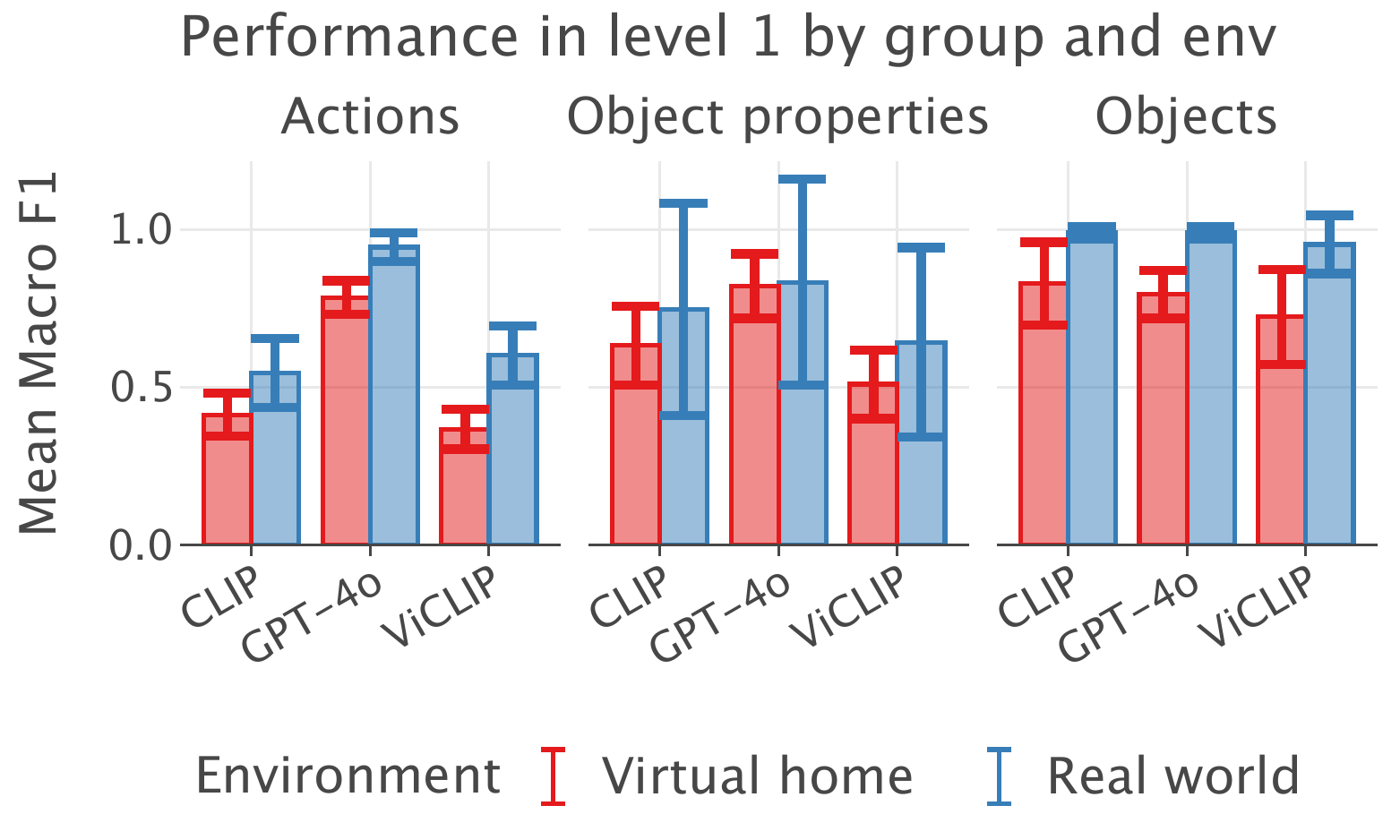}
        \caption{
            \textbf{Real videos are easier than simulations.}
            We see better performance on real world problem sets than on their virtual home counterparts.
            This is especially noticeable on problem sets testing action understanding.
            Additionally, while all the models have strong object recognition capabilities, {\clip} and {\viclip} struggle with actions and object properties.
        }
        \label{fig:results-single-action}
    \end{subfigure}
    \hfill
    \begin{subfigure}[t]{0.49\textwidth}
        \includegraphics[width=\textwidth]{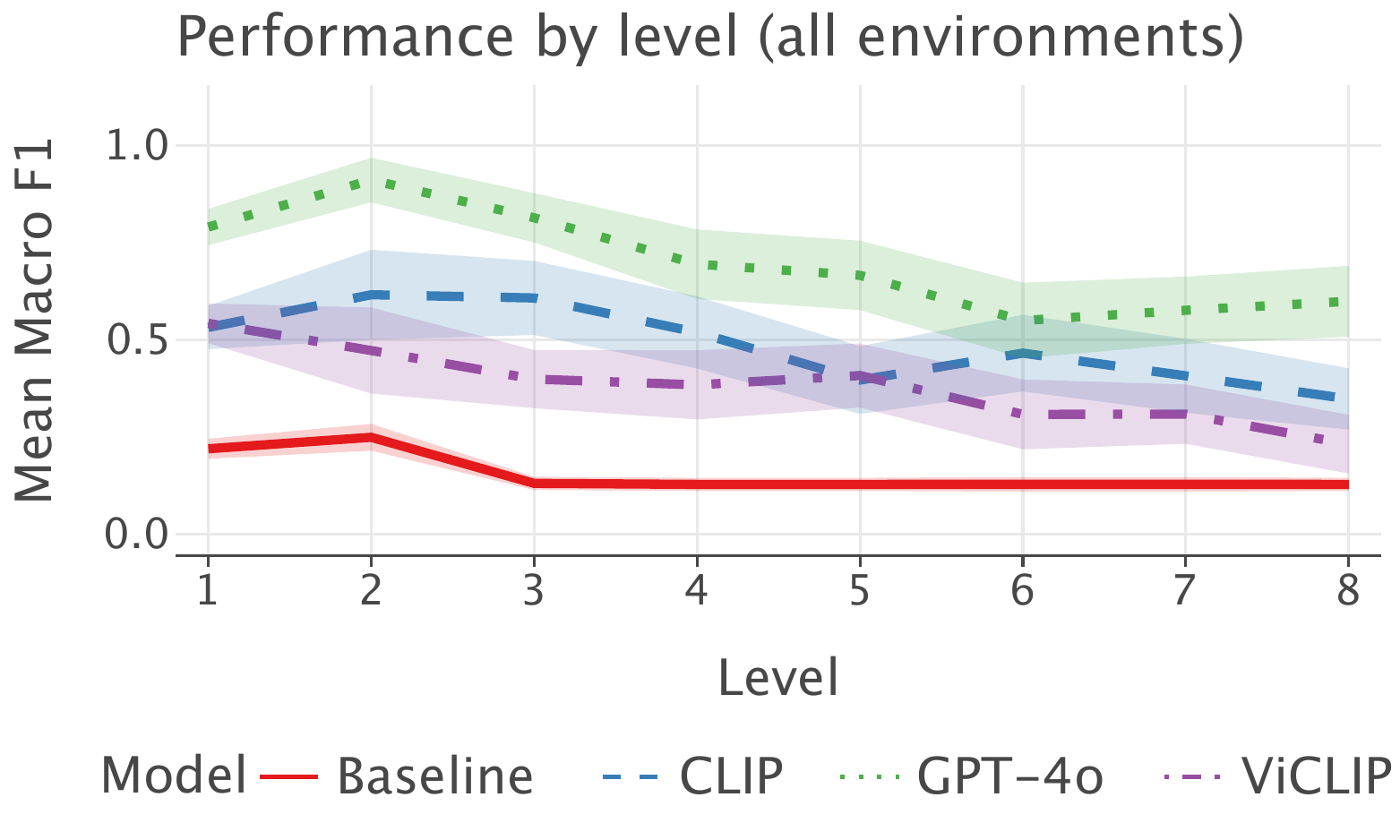}
        \caption{
            \textbf{Longer tasks are more difficult.} 
            The higher up the hierarchy we move, the lower the score (averaged across all three environments). 
            We include a majority class predictor as a baseline.
            The drop levels off for {\gptfour}o, likely because it is possible to match videos with their descriptions using only partial understanding of the action order. We break down the results by task group in \cref{fig:results-multiple-action}, where the drop is more noticeable.
        }\label{fig:results-levels}
    \end{subfigure}
    
    \caption{
        Macro \fscore{} averaged over groups of problem sets in level 1 (\cref{fig:results-single-action}) and in higher levels (\cref{fig:results-levels}).
        Error ranges are 95\% C.I.
    }
\end{figure}

In level 1 tasks (\cref{fig:results-single-action}), we notice several interesting patterns:

\begin{itemize}
    \item \textbf{Real videos are easier than simulations:} Models do less well on tasks in the simulated virtual home environment, compared to analogous tasks in the real world (\cref{fig:results-single-action}, \ref{fig:objects-habitat-vs-real}).
    We hypothesize this is because the simulations are more off-distribution.
    
    \item \textbf{Object recognition is easy:} As expected~\cite{wangPaxionPatchingAction2023}, all the models do well on problem sets which only require object recognition.

    \item \textbf{Recognizing properties is not easy:} Recognizing not only the type of the object, but also its properties---e.g., whether the light is turned on or off---is difficult. Compared to object recognition, we see a drop in performance for {\clip} and {\viclip}.

    \item \textbf{Understanding actions is hard:} Performance of {\clip} and {\viclip} is poor for tasks that require them to check whether a specific action has been performed correctly.
    In \cref{fig:results-full-virtual-home} we see that even {\gptfour}o struggles with complex actions, such as heating things up.

\end{itemize}

\begin{figure}[ht]
    \includegraphics[width=\textwidth]{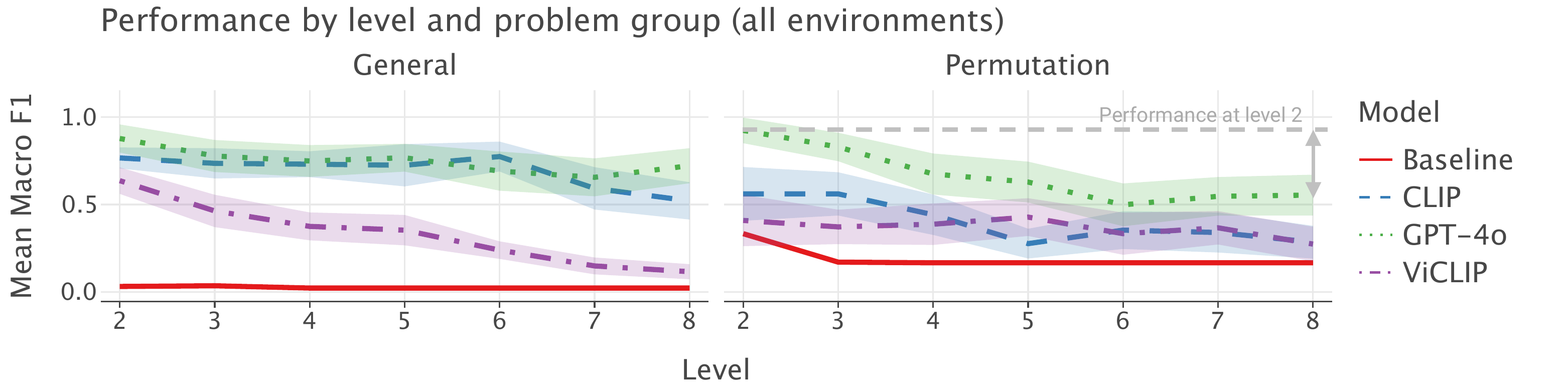}
    
    \caption{
        \textbf{Understanding action order in long videos is hard.}
        In permutation problems, which focus on testing action-order understanding,
        {\gptfour}o's performance starts dropping after level 4, ending up at around 50\% of its original value for videos with 8 actions.
        This is not great performance, considering that the majority class predictor baseline is quite high already.
        The other two models are barely above the baseline.
        Notably, we do not observe this behavior in the general problem sets.
        Despite the baseline being lower there, the models all have higher performance, and --- except for {\viclip} --- retain it from level 2 all the way through level 8.
        Error ranges are 95\% C.I.
    }
    \label{fig:results-multiple-action}
\end{figure}

\begin{figure}[ht]
 \includegraphics[width=\textwidth]{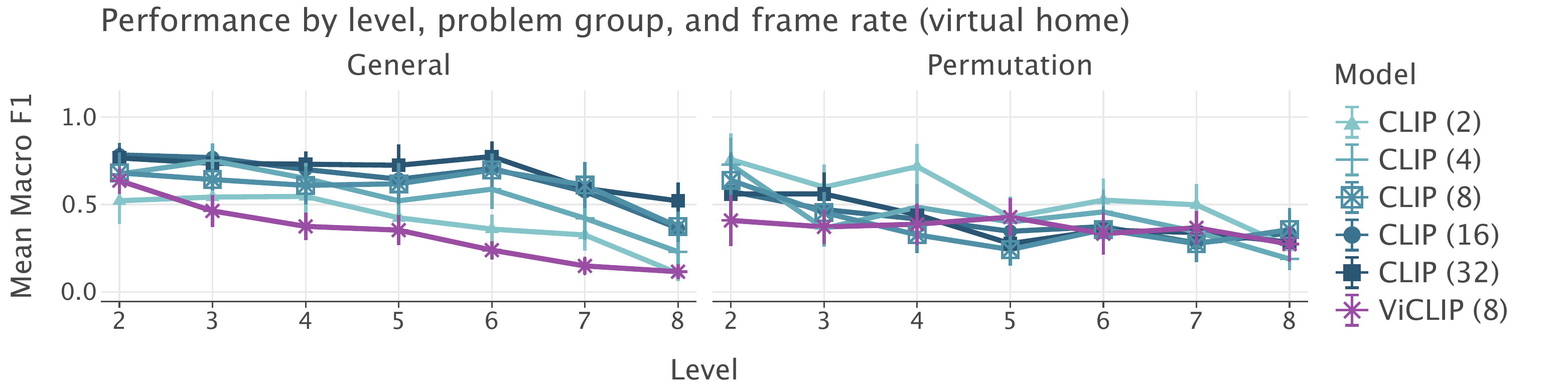}
 \caption{
    \textbf{Frame rate and model scale play an important role in general sequential tasks.}
    We see that the performance of {\clip} rises with increasing frame rate.
    When we compare {\clip}-8 and {\viclip}, which both get 8 frames, we see that {\clip}-8 nevertheless does much better.
    This is likely due to its larger scale, since other differences between the models are minimal.
    Ranges are 95\% C.I.
 }
 \label{fig:results-framerate}
\end{figure}

In multiple-step tasks, we see a gradual drop in performance with rising problem set level (\cref{fig:results-levels}). We find the following:

\begin{itemize}
    \item \textbf{Understanding action order is very hard.} This can most easily be seen when we look at the subset of multiple-action problem sets that test action order understanding (\cref{fig:results-multiple-action}).

    \item \textbf{General sequential tasks can be distinguished without understanding action order.} 
    Videos that do not specifically test for action order understanding can often be matched to their descriptions even by {\clip}, a model that has \emph{no} sense of action order by construction (\cref{fig:results-multiple-action}). 
    We find that in these tasks --- in contrast with the ones testing action order understanding --- frame rate and model size are the most important factors in determining model performance (\cref{fig:results-framerate}).
    This is consistent with previous results showing that {\vlms} rely on their superb object recognition skills even in more complex tasks and can thus easily appear to possess capabilities that they do not have~\cite{wangPaxionPatchingAction2023}.
\end{itemize}

\section{Conclusion}

Useful problems often have sequential structure. 
If {\vlms} are to supervise such problems, they must understand the constituent sub-tasks and their order. 
We present {\capvista}, a hierarchical dataset for testing this understanding.
We illustrate model evaluations with {\vista}, and confirm that {\vlms} are good at object recognition but struggle with sequential tasks.

\bibliographystyle{unsrtnat}
\bibliography{bib}

\begin{thebibliography}{32}
\providecommand{\natexlab}[1]{#1}
\providecommand{\url}[1]{\texttt{#1}}
\expandafter\ifx\csname urlstyle\endcsname\relax
  \providecommand{\doi}[1]{doi: #1}\else
  \providecommand{\doi}{doi: \begingroup \urlstyle{rm}\Url}\fi

\bibitem[Silver et~al.(2016)Silver, Huang, Maddison, Guez, Sifre, Van Den~Driessche, Schrittwieser, Antonoglou, Panneershelvam, Lanctot, et~al.]{alphago}
David Silver, Aja Huang, Chris~J Maddison, Arthur Guez, Laurent Sifre, George Van Den~Driessche, Julian Schrittwieser, Ioannis Antonoglou, Veda Panneershelvam, Marc Lanctot, et~al.
\newblock Mastering the game of go with deep neural networks and tree search.
\newblock \emph{nature}, 529\penalty0 (7587):\penalty0 484--489, 2016.

\bibitem[Tang et~al.(2024)Tang, Abbatematteo, Hu, Chandra, Mart{\'\i}n-Mart{\'\i}n, and Stone]{tang2024deep}
Chen Tang, Ben Abbatematteo, Jiaheng Hu, Rohan Chandra, Roberto Mart{\'\i}n-Mart{\'\i}n, and Peter Stone.
\newblock Deep reinforcement learning for robotics: A survey of real-world successes.
\newblock \emph{arXiv preprint arXiv:2408.03539}, 2024.

\bibitem[Krakovna et~al.(2020)Krakovna, Uesato, Mikulik, Rahtz, Everitt, Kumar, Kenton, Leike, and Legg]{krakovna2020specification}
Victoria Krakovna, Jonathan Uesato, Vladimir Mikulik, Matthew Rahtz, Tom Everitt, Ramana Kumar, Zac Kenton, Jan Leike, and Shane Legg.
\newblock Specification gaming: the flip side of ai ingenuity.
\newblock \emph{DeepMind Blog}, 3, 2020.

\bibitem[Skalse et~al.(2022)Skalse, Howe, Krasheninnikov, and Krueger]{skalse2022defining}
Joar Skalse, Nikolaus Howe, Dmitrii Krasheninnikov, and David Krueger.
\newblock Defining and characterizing reward gaming.
\newblock \emph{Advances in Neural Information Processing Systems}, 35:\penalty0 9460--9471, 2022.

\bibitem[Zhang et~al.(2021)Zhang, Rajan, Pineda, Lambert, Biedenkapp, Chua, Hutter, and Calandra]{halfcheetah}
Baohe Zhang, Raghu Rajan, Luis Pineda, Nathan Lambert, Andr{\'e} Biedenkapp, Kurtland Chua, Frank Hutter, and Roberto Calandra.
\newblock On the importance of hyperparameter optimization for model-based reinforcement learning.
\newblock In \emph{International Conference on Artificial Intelligence and Statistics}, pages 4015--4023. PMLR, 2021.

\bibitem[Fu et~al.(2018)Fu, Singh, Ghosh, Yang, and Levine]{trickvisionclassifier}
Justin Fu, Avi Singh, Dibya Ghosh, Larry Yang, and Sergey Levine.
\newblock Variational inverse control with events: A general framework for data-driven reward definition.
\newblock \emph{Advances in neural information processing systems}, 31, 2018.

\bibitem[Christiano et~al.(2017)Christiano, Leike, Brown, Martic, Legg, and Amodei]{backflipandtrickyrobothand}
Paul~F Christiano, Jan Leike, Tom Brown, Miljan Martic, Shane Legg, and Dario Amodei.
\newblock Deep reinforcement learning from human preferences.
\newblock \emph{Advances in neural information processing systems}, 30, 2017.

\bibitem[Rocamonde et~al.(2023)Rocamonde, Montesinos, Nava, Perez, and Lindner]{rocamondeVisionLanguageModelsAre2023}
Juan Rocamonde, Victoriano Montesinos, Elvis Nava, Ethan Perez, and David Lindner.
\newblock Vision-{{Language Models}} are {{Zero-Shot Reward Models}} for {{Reinforcement Learning}}, 2023.
\newblock URL \url{http://arxiv.org/abs/2310.12921}.

\bibitem[Sontakke et~al.(2023)Sontakke, Zhang, Arnold, Pertsch, Bıyık, Sadigh, Finn, and Itti]{sontakkeRoboCLIPOneDemonstration2023}
Sumedh~A. Sontakke, Jesse Zhang, Sébastien M.~R. Arnold, Karl Pertsch, Erdem Bıyık, Dorsa Sadigh, Chelsea Finn, and Laurent Itti.
\newblock {{RoboCLIP}}: {{One Demonstration}} is {{Enough}} to {{Learn Robot Policies}}, 2023.
\newblock URL \url{http://arxiv.org/abs/2310.07899}.

\bibitem[Guan et~al.(2024)Guan, Zhou, Liu, Zha, Amor, and Kambhampati]{guanTaskSuccessNot2024}
Lin Guan, Yifan Zhou, Denis Liu, Yantian Zha, Heni~Ben Amor, and Subbarao Kambhampati.
\newblock "{{Task Success}}" is not {{Enough}}: {{Investigating}} the {{Use}} of {{Video-Language Models}} as {{Behavior Critics}} for {{Catching Undesirable Agent Behaviors}}, 2024.
\newblock URL \url{http://arxiv.org/abs/2402.04210}.

\bibitem[Chen et~al.(2021)Chen, Nair, and Finn]{inthewildvideos}
Annie~S Chen, Suraj Nair, and Chelsea Finn.
\newblock Learning generalizable robotic reward functions from" in-the-wild" human videos.
\newblock \emph{arXiv preprint arXiv:2103.16817}, 2021.

\bibitem[Parisi et~al.(2022)Parisi, Rajeswaran, Purushwalkam, and Gupta]{manypretrainedinclclip}
Simone Parisi, Aravind Rajeswaran, Senthil Purushwalkam, and Abhinav Gupta.
\newblock The unsurprising effectiveness of pre-trained vision models for control.
\newblock In \emph{international conference on machine learning}, pages 17359--17371. PMLR, 2022.

\bibitem[Kwon et~al.(2023)Kwon, Xie, Bullard, and Sadigh]{llmsupervisornovisual}
Minae Kwon, Sang~Michael Xie, Kalesha Bullard, and Dorsa Sadigh.
\newblock Reward design with language models.
\newblock \emph{arXiv preprint arXiv:2303.00001}, 2023.

\bibitem[Lambert et~al.(2024)Lambert, Pyatkin, Morrison, Miranda, Lin, Chandu, Dziri, Kumar, Zick, Choi, Smith, and Hajishirzi]{lambert2024rewardbench}
Nathan Lambert, Valentina Pyatkin, Jacob Morrison, LJ~Miranda, Bill~Yuchen Lin, Khyathi Chandu, Nouha Dziri, Sachin Kumar, Tom Zick, Yejin Choi, Noah~A. Smith, and Hannaneh Hajishirzi.
\newblock Rewardbench: Evaluating reward models for language modeling.
\newblock \emph{CoRR}, abs/2403.13787, 2024.
\newblock \doi{10.48550/ARXIV.2403.13787}.
\newblock URL \url{https://doi.org/10.48550/arXiv.2403.13787}.

\bibitem[Tung et~al.(2018)Tung, Harley, Huang, and Fragkiadaki]{languageandphotodemo}
Hsiao-Yu Tung, Adam~W Harley, Liang-Kang Huang, and Katerina Fragkiadaki.
\newblock Reward learning from narrated demonstrations.
\newblock In \emph{Proceedings of the IEEE Conference on Computer Vision and Pattern Recognition}, pages 7004--7013, 2018.

\bibitem[Mahmoudieh et~al.(2022)Mahmoudieh, Pathak, and Darrell]{clipsupervisor}
Parsa Mahmoudieh, Deepak Pathak, and Trevor Darrell.
\newblock Zero-shot reward specification via grounded natural language.
\newblock In \emph{International Conference on Machine Learning}, pages 14743--14752. PMLR, 2022.

\bibitem[Wang et~al.(2024{\natexlab{a}})Wang, Sun, Zhang, Xian, Biyik, Held, and Erickson]{wangRLVLMFReinforcementLearning2024}
Yufei Wang, Zhanyi Sun, Jesse Zhang, Zhou Xian, Erdem Biyik, David Held, and Zackory Erickson.
\newblock {{RL-VLM-F}}: {{Reinforcement Learning}} from {{Vision Language Foundation Model Feedback}}, 2024{\natexlab{a}}.
\newblock URL \url{http://arxiv.org/abs/2402.03681}.

\bibitem[Chen et~al.(2024)Chen, Zhang, Ren, Zhao, Cai, Wang, Wang, Meng, Liu, and Chang]{chenPCABenchEvaluatingMultimodal2024}
Liang Chen, Yichi Zhang, Shuhuai Ren, Haozhe Zhao, Zefan Cai, Yuchi Wang, Peiyi Wang, Xiangdi Meng, Tianyu Liu, and Baobao Chang.
\newblock {{PCA-Bench}}: {{Evaluating Multimodal Large Language Models}} in {{Perception-Cognition-Action Chain}}, 2024.
\newblock URL \url{http://arxiv.org/abs/2402.15527}.

\bibitem[Wang et~al.(2023)Wang, Blume, Li, Liu, Cho, Tang, Bansal, and Ji]{wangPaxionPatchingAction2023}
Zhenhailong Wang, Ansel Blume, Sha Li, Genglin Liu, Jaemin Cho, Zineng Tang, Mohit Bansal, and Heng Ji.
\newblock Paxion: {{Patching Action Knowledge}} in {{Video-Language Foundation Models}}, 2023.
\newblock URL \url{http://arxiv.org/abs/2305.10683}.

\bibitem[Li et~al.(2024)Li, Zhang, Wong, Gokmen, Srivastava, Martín-Martín, Wang, Levine, Ai, Martinez, Yin, Lingelbach, Hwang, Hiranaka, Garlanka, Aydin, Lee, Sun, Anvari, Sharma, Bansal, Hunter, Kim, Lou, Matthews, Villa-Renteria, Tang, Tang, Xia, Li, Savarese, Gweon, Liu, Wu, and Fei-Fei]{liBEHAVIOR1KHumanCenteredEmbodied2024}
Chengshu Li, Ruohan Zhang, Josiah Wong, Cem Gokmen, Sanjana Srivastava, Roberto Martín-Martín, Chen Wang, Gabrael Levine, Wensi Ai, Benjamin Martinez, Hang Yin, Michael Lingelbach, Minjune Hwang, Ayano Hiranaka, Sujay Garlanka, Arman Aydin, Sharon Lee, Jiankai Sun, Mona Anvari, Manasi Sharma, Dhruva Bansal, Samuel Hunter, Kyu-Young Kim, Alan Lou, Caleb~R. Matthews, Ivan Villa-Renteria, Jerry~Huayang Tang, Claire Tang, Fei Xia, Yunzhu Li, Silvio Savarese, Hyowon Gweon, C.~Karen Liu, Jiajun Wu, and Li~Fei-Fei.
\newblock {{BEHAVIOR-1K}}: {{A Human-Centered}}, {{Embodied AI Benchmark}} with 1,000 {{Everyday Activities}} and {{Realistic Simulation}}, 2024.
\newblock URL \url{http://arxiv.org/abs/2403.09227}.

\bibitem[Puig et~al.(2018)Puig, Ra, Boben, Li, Wang, Fidler, and Torralba]{puigVirtualHomeSimulatingHousehold2018}
Xavier Puig, Kevin Ra, Marko Boben, Jiaman Li, Tingwu Wang, Sanja Fidler, and Antonio Torralba.
\newblock {{VirtualHome}}: {{Simulating Household Activities}} via {{Programs}}, 2018.
\newblock URL \url{http://arxiv.org/abs/1806.07011}.

\bibitem[Savva et~al.(2019)Savva, Kadian, Maksymets, Zhao, Wijmans, Jain, Straub, Liu, Koltun, Malik, Parikh, and Batra]{savvaHabitatPlatformEmbodied2019}
Manolis Savva, Abhishek Kadian, Oleksandr Maksymets, Yili Zhao, Erik Wijmans, Bhavana Jain, Julian Straub, Jia Liu, Vladlen Koltun, Jitendra Malik, Devi Parikh, and Dhruv Batra.
\newblock Habitat: {{A Platform}} for {{Embodied AI Research}}, 2019.
\newblock URL \url{http://arxiv.org/abs/1904.01201}.

\bibitem[Puig et~al.(2023)Puig, Undersander, Szot, Cote, Yang, Partsey, Desai, Clegg, Hlavac, Min, Vondruš, Gervet, Berges, Turner, Maksymets, Kira, Kalakrishnan, Malik, Chaplot, Jain, Batra, Rai, and Mottaghi]{puigHabitatCoHabitatHumans2023a}
Xavier Puig, Eric Undersander, Andrew Szot, Mikael~Dallaire Cote, Tsung-Yen Yang, Ruslan Partsey, Ruta Desai, Alexander~William Clegg, Michal Hlavac, So~Yeon Min, Vladimír Vondruš, Theophile Gervet, Vincent-Pierre Berges, John~M. Turner, Oleksandr Maksymets, Zsolt Kira, Mrinal Kalakrishnan, Jitendra Malik, Devendra~Singh Chaplot, Unnat Jain, Dhruv Batra, Akshara Rai, and Roozbeh Mottaghi.
\newblock Habitat 3.0: {{A Co-Habitat}} for {{Humans}}, {{Avatars}} and {{Robots}}, 2023.
\newblock URL \url{http://arxiv.org/abs/2310.13724}.

\bibitem[Shridhar et~al.(2020)Shridhar, Thomason, Gordon, Bisk, Han, Mottaghi, Zettlemoyer, and Fox]{shridharALFREDBenchmarkInterpreting2020}
Mohit Shridhar, Jesse Thomason, Daniel Gordon, Yonatan Bisk, Winson Han, Roozbeh Mottaghi, Luke Zettlemoyer, and Dieter Fox.
\newblock {{ALFRED}}: {{A Benchmark}} for {{Interpreting Grounded Instructions}} for {{Everyday Tasks}}, 2020.
\newblock URL \url{http://arxiv.org/abs/1912.01734}.

\bibitem[Carreira et~al.(2019)Carreira, Noland, Hillier, and Zisserman]{kinetics}
Joao Carreira, Eric Noland, Chloe Hillier, and Andrew Zisserman.
\newblock A short note on the kinetics-700 human action dataset.
\newblock \emph{arXiv preprint arXiv:1907.06987}, 2019.

\bibitem[Milani et~al.(2023)Milani, Kanervisto, Ramanauskas, Schulhoff, Houghton, and Shah]{milani2023bedd}
Stephanie Milani, Anssi Kanervisto, Karolis Ramanauskas, Sander Schulhoff, Brandon Houghton, and Rohin Shah.
\newblock {BEDD:} the minerl {BASALT} evaluation and demonstrations dataset for training and benchmarking agents that solve fuzzy tasks.
\newblock In Alice Oh, Tristan Naumann, Amir Globerson, Kate Saenko, Moritz Hardt, and Sergey Levine, editors, \emph{Advances in Neural Information Processing Systems 36: Annual Conference on Neural Information Processing Systems 2023, NeurIPS 2023, New Orleans, LA, USA, December 10 - 16, 2023}, 2023.
\newblock URL \url{http://papers.nips.cc/paper\_files/paper/2023/hash/67a6726dcd555b982cabb3446ffac01d-Abstract-Datasets\_and\_Benchmarks.html}.

\bibitem[Yuan et~al.(2023)Yuan, Zhang, Wang, Xie, Cai, Dong, and Lu]{yuan2023plan4mc}
Haoqi Yuan, Chi Zhang, Hongcheng Wang, Feiyang Xie, Penglin Cai, Hao Dong, and Zongqing Lu.
\newblock Plan4mc: Skill reinforcement learning and planning for open-world minecraft tasks.
\newblock \emph{CoRR}, abs/2303.16563, 2023.
\newblock \doi{10.48550/ARXIV.2303.16563}.
\newblock URL \url{https://doi.org/10.48550/arXiv.2303.16563}.

\bibitem[Radford et~al.(2021)Radford, Kim, Hallacy, Ramesh, Goh, Agarwal, Sastry, Askell, Mishkin, Clark, et~al.]{clip}
Alec Radford, Jong~Wook Kim, Chris Hallacy, Aditya Ramesh, Gabriel Goh, Sandhini Agarwal, Girish Sastry, Amanda Askell, Pamela Mishkin, Jack Clark, et~al.
\newblock Learning transferable visual models from natural language supervision.
\newblock In \emph{International conference on machine learning}, pages 8748--8763. PMLR, 2021.

\bibitem[Cherti et~al.(2022)Cherti, Beaumont, Wightman, Wortsman, Ilharco, Gordon, Schuhmann, Schmidt, and Jitsev]{chertiReproducibleScalingLaws2022}
Mehdi Cherti, Romain Beaumont, Ross Wightman, Mitchell Wortsman, Gabriel Ilharco, Cade Gordon, Christoph Schuhmann, Ludwig Schmidt, and Jenia Jitsev.
\newblock Reproducible scaling laws for contrastive language-image learning, 2022.
\newblock URL \url{http://arxiv.org/abs/2212.07143}.

\bibitem[Schuhmann et~al.(2022)Schuhmann, Beaumont, Vencu, Gordon, Wightman, Cherti, Coombes, Katta, Mullis, Wortsman, Schramowski, Kundurthy, Crowson, Schmidt, Kaczmarczyk, and Jitsev]{schuhmannLAION5BOpenLargescale2022}
Christoph Schuhmann, Romain Beaumont, Richard Vencu, Cade Gordon, Ross Wightman, Mehdi Cherti, Theo Coombes, Aarush Katta, Clayton Mullis, Mitchell Wortsman, Patrick Schramowski, Srivatsa Kundurthy, Katherine Crowson, Ludwig Schmidt, Robert Kaczmarczyk, and Jenia Jitsev.
\newblock {{LAION-5B}}: {{An}} open large-scale dataset for training next generation image-text models, 2022.
\newblock URL \url{http://arxiv.org/abs/2210.08402}.

\bibitem[Wang et~al.(2024{\natexlab{b}})Wang, He, Li, Li, Yu, Ma, Li, Chen, Chen, Wang, He, Luo, Liu, Wang, Wang, and Qiao]{viclip}
Yi~Wang, Yinan He, Yizhuo Li, Kunchang Li, Jiashuo Yu, Xin Ma, Xinhao Li, Guo Chen, Xinyuan Chen, Yaohui Wang, Conghui He, Ping Luo, Ziwei Liu, Yali Wang, Limin Wang, and Yu~Qiao.
\newblock {{InternVid}}: {{A Large-scale Video-Text Dataset}} for {{Multimodal Understanding}} and {{Generation}}, 2024{\natexlab{b}}.
\newblock URL \url{http://arxiv.org/abs/2307.06942}.

\bibitem[Hurst et~al.(2024)Hurst, Lerer, Goucher, Perelman, Ramesh, Clark, Ostrow, Welihinda, Hayes, Radford, et~al.]{hurst2024gpt}
Aaron Hurst, Adam Lerer, Adam~P Goucher, Adam Perelman, Aditya Ramesh, Aidan Clark, AJ~Ostrow, Akila Welihinda, Alan Hayes, Alec Radford, et~al.
\newblock Gpt-4o system card.
\newblock \emph{arXiv preprint arXiv:2410.21276}, 2024.

\end{thebibliography}


\clearpage

\section*{Author contribution statement} \label{app:contributions}

Evžen Wybitul:
\begin{itemize}
    \item Implemented the current evaluation repository.
    \item Was one of the sources for the idea of building a structured, hierarchical dataset.
    \item Wrote most of the sections in the original and current versions of the paper.
    \item For an earlier version of the paper, manually recorded hundreds of videos in a virtual home environment. For the current version, wrote code that re-mixes the ALFRED dataset, producing 3,000+ videos.
    \item Helped plan out the work and coordinate the team.
\end{itemize}

Mikhail Seleznyov:

\begin{itemize}
    \item Implemented the initial repository for first experiments with trajectory evaluation.
    \item Made a limited set of high quality trajectories.
    \item Extended the tests from virtual home to Minecraft.
    \item Implemented first quantitative metrics and visualizations of model's performance.
    \item Helped Evžen with conceptualization of task hierarchy for first iterations of the paper.
    \item Helped Evžen with writing the first version of the paper.
\end{itemize}

Evan Ryan Gunter:

\begin{itemize}
    \item With Evžen, decided which real world videos to create to adequately compare with virtual home; created/adapted the task descriptions and model prompts for the real world videos.
    \item Created, processed, and labeled the 923 real world videos in the virtual home recreation, object tracking, and object interaction categories, and manually ensured that they contain sufficient information for a human to achieve 100\% accuracy on all tasks.
    \item Sourced the 200 complex-environment action recognition videos from kinetics-700.
    \item Evaluated all the real world videos, using the same methodology as in virtual home for the virtual home remake videos, and extended this methodology to the other categories of real world video with no analogues in virtual home or Minecraft.
    \item Wrote code to check data integrity for the real world data.
    \item Helped with writing and editing the current version of the paper, and significantly extended the literature review.
    \item Strategy discussions in meetings and individually with Evžen.
\end{itemize}

\appendix

\section{Appendix}

\subsection{Dataset construction and its structure} \label{app:problem-sets}

In this section, we detail the structure of the dataset, the construction of the problem sets, and the way we produced the videos in the different environments.

\subsubsection{Problem sets}

The number of videos and descriptions in a problem set varies between environments and problem types. In general, though, the following holds: 

\begin{itemize}
    \item Level 1 problem sets have anywhere from around 10 to around 300 videos. Often, there will be multiple (usually 10) videos sharing a description for added robustness.
    \item Permutation problem sets have three videos, each with a different description. There are multiple permutation problem sets per level.
    \item The general multiple-step problem sets have 9 videos. Depending on the level, a remix task can have multiple classes per prefix length --- on level 2, for example, it will have 4 videos that share the first step with the base video, and 4 videos that are different from the very beginning. Each video has a unique description.
\end{itemize}

\paragraph{General multiple-step problem sets.} These problem sets contain 9 videos: 1 base video and 8 videos derived from the base one. The derived videos share prefixes of different lengths with the base video. This includes trajectories that have a shared length of 0, meaning they are completely different from the base. This simulates the scenario of the model supervising trajectories with different amounts of similarity between each other and having to notice the differences.

\paragraph{Permutation problem sets.} These problem sets contain 3 videos depicting different permutations of the same steps. The permutations in any given set are chosen randomly, though we do ensure that the step combinations are logically consistent (e.g. no putting down an object before picking it up).

\subsubsection{Virtual home}

Virtual home videos comprise a majority of the dataset. The videos were automatically generated from the {\alfred} \cite{shridharALFREDBenchmarkInterpreting2020} dataset.

\paragraph{Level 1, single-action videos.} We generated the level 1 videos by extracting actions and objects from videos in the long-horizon tasks in {\alfred} \cite{shridharALFREDBenchmarkInterpreting2020}. 
After organizing the videos into problem sets, we have got 2 groups of problem sets related to object recognition, 2 groups related to recognizing different object states (both types are listed in \cref{tab:object-problems}), and 6 groups related to understanding different actions (\cref{tab:list-action-problems}).

\paragraph{Problems in higher difficulty levels} In virtual home, the videos in higher levels are generated automatically by stitching together short clips of different {\alfred} videos. This can result in small visual glitches, such as sudden teleportation, changes in the objects the agent is holding, or minor inconsistencies in the general background scene. These glitches are not problematic for our evaluations, since the models all have low frame-rates. Nevertheless, we make effort to minimize these: we make sure to only chain actions that can logically follow each other (e.g. we can only put down a thing we were holding in the previous frames), and only from sources that take place in the same type of room. You can see a list of the high-level problem sets in \cref{tab:list-high-level-problems}.

\begin{table}[ht]
\centering
\begin{tabular}{llll}
\toprule
Level & Group          & Task                   & Classes, videos per problem                 \\
\midrule
Base  & Heat           & Bread                  & 5 classes, 50 videos   \\
Base  & Heat           & Slice of Potato        & 5 classes, 50 videos   \\
Base  & Heat           & Potato                 & 5 classes, 50 videos   \\
Base  & Heat           & Mug                    & 5 classes, 50 videos   \\
Base  & Heat           & Apple                  & 5 classes, 50 videos   \\
Base  & Slice          & Bread                  & 2 classes, 20 videos   \\
Base  & Slice          & Lettuce                & 2 classes, 20 videos   \\
Base  & Slice          & Tomato                 & 2 classes, 20 videos   \\
Base  & Slice          & Potato                 & 2 classes, 20 videos   \\
Base  & Slice          & Apple                  & 2 classes, 20 videos   \\
Base  & Pick vs Put    & Pen Garbage Can        & 2 classes, 11 videos   \\
Base  & Pick vs Put    & Egg Microwave          & 2 classes, 20 videos   \\
Base  & Pick vs Put    & Cell Phone Side Table  & 2 classes, 19 videos   \\
Base  & Pick vs Put    & Fork Counter Top       & 2 classes, 15 videos   \\
Base  & Pick vs Put    & Salt Shaker Cabinet    & 2 classes, 20 videos   \\
Base  & Toggle         & Faucet                 & 2 classes, 20 videos   \\
Base  & Toggle         & Desk Lamp              & 2 classes, 20 videos   \\
Base  & Toggle         & Floor Lamp             & 2 classes, 20 videos   \\
Base  & Toggle         & Microwave              & 2 classes, 20 videos   \\
Base  & Clean          & Cloth                  & 3 classes, 30 videos   \\
Base  & Clean          & Kettle                 & 3 classes, 30 videos   \\
Base  & Clean          & Apple                  & 3 classes, 30 videos   \\
Base  & Clean          & Egg                    & 3 classes, 30 videos   \\
Base  & Clean          & Fork                   & 3 classes, 30 videos   \\
Base  & Cool           & Slice of Tomato        & 4 classes, 40 videos   \\
Base  & Cool           & Lettuce                & 4 classes, 40 videos   \\
Base  & Cool           & Tomato                 & 4 classes, 40 videos   \\
Base  & Cool           & Egg                    & 4 classes, 40 videos   \\
Base  & Cool           & Slice of Bread         & 4 classes, 40 videos   \\
\bottomrule
\end{tabular}
\vspace{0.3cm}
\caption{Number of classes and videos per each level 1 problem set that tests action understanding in the virtual home environment.}
\label{tab:list-action-problems}
\end{table}

\begin{table}[ht]
\centering
\begin{tabular}{llll}
\toprule
Level & Group          & Task                   & Classes, videos per problem                 \\
\midrule
Base  & Objects        & Pick from Dining Table & 12 classes, 88 videos  \\
Base  & Objects        & Pick from Counter Top  & 40 classes, 364 videos \\
Base  & Objects        & Pick from Somewhere    & 5 classes, 50 videos   \\
Base  & Containers     & Place Key Chain        & 5 classes, 49 videos   \\
Base  & Containers     & Place Butter Knife     & 10 classes, 91 videos  \\
Base  & Containers     & Place Mug              & 3 classes, 22 videos   \\
Base  & Containers     & Place Soap Bar         & 3 classes, 30 videos   \\
Base  & Sliced vs Whole& Bread                  & 2 classes, 20 videos   \\
Base  & Sliced vs Whole& Lettuce                & 2 classes, 20 videos   \\
Base  & Sliced vs Whole& Tomato                 & 2 classes, 20 videos   \\
Base  & Sliced vs Whole& Potato                 & 2 classes, 20 videos   \\
Base  & Sliced vs Whole& Apple                  & 2 classes, 20 videos   \\
Base  & On vs Off      & Faucet                 & 2 classes, 20 videos   \\
Base  & On vs Off      & Desk Lamp              & 2 classes, 20 videos   \\
Base  & On vs Off      & Floor Lamp             & 2 classes, 20 videos   \\
Base  & On vs Off      & Microwave              & 2 classes, 20 videos   \\
\bottomrule
\end{tabular}
\vspace{0.3cm}
\caption{Number of classes and videos per each level 1 problem set that tests object and object property understanding in the virtual home environment.}
\label{tab:object-problems}
\end{table}

\begin{table}[ht]
\centering
\begin{tabular}{llll}
\toprule
Level & Group          & Task                   & Classes, videos per problem                 \\
\midrule
Level 2–8  & Remix     & (12 problems per level)         & 8 classes, 8 videos   \\
Level 2  & Permutation     & (32 problems per level)         & 2 classes, 2 videos   \\
Level 3–8  & Permutation     & (33 problems per level)         & 3 classes, 3 videos   \\
\bottomrule
\end{tabular}
\vspace{0.3cm}
\caption{Number of classes and videos per each multiple-action problem in the virtual home environment}
\label{tab:list-high-level-problems}
\end{table}

\subsubsection{Minecraft}

The length of the videos spans from 3 seconds to slightly over 1 minute.
In total this part includes 32 videos for single-action problems and 21 videos for multiple-action problems.

\paragraph{Level 1, single-action videos.} Inspired by the skill categorization proposed by \citet{yuan2023plan4mc}, we identify 7 fundamental actions in Minecraft: placing blocks, breaking blocks, crafting, combat, finding something, mining stone-like blocks with a pickax, and picking up an item which was dropped after breaking a block or defeating an enemy or animal,

Mining is a strict subset of breaking blocks, but we decided to add this action as it is especially important in Minecraft gameplay.
We restrict the finding skill to finding animals (we consider a video an example of ``finding'' if in the first frame there is no animal, and then it appears in a later frame).

\paragraph{Multiple-action problems.} We generate plausible trajectories, which are sequences of the fundamental actions. A typical pattern is ``break-craft'' (e.g. to make sticks after breaking a tree), or ``find-combat-pick up'' (notice an animal, attack it, and gather the dropped items).

\subsubsection{Real world}

There are four types of videos in the real world dataset: virtual home mimic videos, object tracking videos, object-interaction videos, and non-controlled-environment videos.

\paragraph{Virtual home mimic videos.}
These videos are intended to test similar scenarios to the virtual home videos, to allow for comparison between simulated and real environments.
The majority of the videos are of the same general format as the {\alfred} videos, with the same or very similar tasks in the same hierarchy; each was created manually in a single take.
There are also a handful of videos that are close recreations of the setting and object movements of specific virtual home videos, to set a baseline for direct comparison.

\paragraph{Object tracking videos.}
One of the more nuanced kinds of understanding that is not well tested by our videos in simulated environments is object tracking---identifying the path of each object throughout the video.
To test this, we recorded videos in which several similar objects are arrayed in a line on a table, moved around for a few seconds, and rearranged into a line.
The model is then asked whether the object that was originally leftmost is still leftmost at the end of the video.
The videos vary in number of objects, how similar the objects are to each other, and how the objects are manipulated.
The object tracking videos also include a few process validation videos: some where the objects being moved around are completely different, and some where objects are picked up from the line without scrambling to test whether the model can identify the leftmost object accurately.

\paragraph{Object interaction videos.}
Like object tracking, our virtual home videos do not thoroughly test understanding of complex interactions between objects.
Objects in the virtual home may have different states, but these states are quite limited---for example, an apple can be cut, but when it is cut it instantly transforms from the whole apple into apple slices.
In the virtual home, there are no ``soft'' objects---objects that can be continuously deformed.
To test understanding of these properties, we include real videos of either pinning two pieces of fabric together, or interacting with fabric and pins in such a way that the final result is that the pieces of fabric are not pinned together (because the pin goes under the fabric instead of through it, or only goes through one piece of fabric, or goes through both pieces of fabric but then retracts, etc.).
Additionally, these videos are somewhat off-distribution compared to most video datasets---they are taken at high magnification and are cropped closely.
This provides a test of how models perform in off-distribution, non-simulated environments.

\paragraph{Non-controlled environment videos.}
Although {\vlms}' performance may suffer due to being off-distribution in our virtual environments, these virtual environment do have advantages over real environments: they are much simpler.
Our videos intended to be closely analogous to the virtual home videos also have a simple, highly controlled environment; so, they may not be a good test of models' ability to understand key actions in a high-complexity environment.
Our non-controlled environment videos aim to test basic action recognition in high-complexity environments.
In particular, these videos consist of 100 videos of a door opening and 100 videos of a door closing from the Kinetics-700~\cite{kinetics} dataset; the task is simply to identify whether the door opened or closed.
These videos were manually selected to contain either a door opening or closing, but not both.

\begin{table}[ht]
\centering
\begin{threeparttable}
\begin{tabular}{llll}
\toprule
Level & Group           & Task                   & Classes, videos per problem                 \\
\midrule
Base  & Heat            & Apple                   &  5 classes, 40 videos   \\
Base  & Heat            & Mug                     &  5 classes, 40 videos   \\
Base  & Slice           & Apple                   &  2 classes,  8 videos   \\
Base  & Slice           & Potato                  &  2 classes,  8 videos   \\
Base  & Pick vs Put     & Apple Counter Top       &  2 classes, 12 videos\tnote{a} \\
Base  & Pick vs Put     & Apple Freezer           &  2 classes, 12 videos\tnote{a}   \\
Base  & Pick vs Put     & Apple Microwave         &  2 classes, 12 videos\tnote{a}   \\
Base  & Pick vs Put     & Apple Sink              &  2 classes, 12 videos\tnote{a}  \\
Base  & Pick vs Put     & Butter Knife Countertop &  2 classes, 12 videos\tnote{b}  \\
Base  & Pick vs Put     & Butter Knife Freezer    &  2 classes, 12 videos\tnote{b}   \\
Base  & Pick vs Put     & Butter Knife Microwave  &  2 classes, 12 videos\tnote{b}   \\
Base  & Pick vs Put     & Butter Knife Sink       &  2 classes, 12 videos\tnote{b}   \\
Base  & Pick vs Put     & Can Counter Top         &  2 classes,  12 videos\tnote{c}   \\
Base  & Pick vs Put     & Can Freezer             &  2 classes,  12 videos\tnote{c}   \\
Base  & Pick vs Put     & Can Microwave           &  2 classes,  12 videos\tnote{c}   \\
Base  & Pick vs Put     & Can Sink                &  2 classes,  12 videos\tnote{c}   \\
Base  & Pick vs Put     & Hammer Counter Top      &  2 classes,  12 videos\tnote{d}   \\
Base  & Pick vs Put     & Hammer Freezer          &  2 classes,  12 videos\tnote{d}   \\
Base  & Pick vs Put     & Hammer Microwave        &  2 classes,  12 videos\tnote{d}   \\
Base  & Pick vs Put     & Hammer Sink             &  2 classes,  12 videos\tnote{d}   \\
Base  & Pick vs Put     & Mug Counter Top         &  2 classes,  12 videos\tnote{e}   \\
Base  & Pick vs Put     & Mug Freezer             &  2 classes,  12 videos\tnote{e}   \\
Base  & Pick vs Put     & Mug Microwave           &  2 classes,  12 videos\tnote{e}   \\
Base  & Pick vs Put     & Mug Sink                &  2 classes,  12 videos\tnote{e}   \\
Base  & Pick vs Put     & Fork Counter Top        &  2 classes,  8 videos   \\
Base  & Toggle          & Desk Lamp               &  2 classes,  2 videos   \\
Base  & Toggle          & Faucet                  &  2 classes,  2 videos   \\
Base  & Toggle          & Floor Lamp              &  2 classes,  6 videos   \\
Base  & Toggle          & Microwave               &  2 classes,  2 videos   \\
Base  & Clean           & Apple                   &  2 classes, 16 videos   \\
Base  & Clean           & Mug                     &  2 classes, 16 videos   \\
Base  & Cool            & Apple                   &  4 classes, 32 videos   \\
Base  & Cool            & Mug                     &  4 classes, 32 videos   \\
\bottomrule
\end{tabular}
    \begin{tablenotes}
      \item[a] 6 shared with \hyperref[tab:object-problems-real-world]{Place Apple}
      \item[b] 6 shared with \hyperref[tab:object-problems-real-world]{Place Butter Knife}
      \item[c] 6 shared with \hyperref[tab:object-problems-real-world]{Place Can}
      \item[d] 6 shared with \hyperref[tab:object-problems-real-world]{Place Hammer}
      \item[e] 6 shared with \hyperref[tab:object-problems-real-world]{Place Mug}
    \end{tablenotes}
  \end{threeparttable}
\vspace{0.3cm}
\caption{Number of classes and videos per each level 1 problem set testing action understanding in the real world environment.}
\label{tab:action-problems-real-world}
\end{table}

\begin{table}[ht]
\centering
\begin{threeparttable}
\begin{tabular}{llll}
\toprule
Level & Group           & Task                    & Classes, videos per problem \\
\midrule
Base  & Objects         & Pick from Counter Top   & 18 classes, 70 videos   \\
Base  & Objects         & Pick from Table         &  6 classes, 20 videos   \\
Base  & Containers      & Place Apple             &  4 classes, 24 videos\tnote{a} \\
Base  & Containers      & Place Butter Knife      &  4 classes, 24 videos\tnote{b} \\
Base  & Containers      & Place Can               &  4 classes, 24 videos\tnote{c} \\
Base  & Containers      & Place Hammer            &  4 classes, 24 videos\tnote{d}   \\
Base  & Containers      & Place Mug               &  4 classes, 24 videos\tnote{e}   \\
Base  & Sliced vs Whole & Apple                   &  2 classes,  8 videos\tnote{f}   \\
Base  & Sliced vs Whole & Potato                  &  2 classes,  16 videos\tnote{g}  \\
Base  & On vs Off       & Desk Lamp               &  2 classes,  2 videos   \\
Base  & On vs Off       & Faucet                  &  2 classes,  2 videos   \\
Base  & On vs Off       & Floor Lamp              &  2 classes,  6 videos   \\
Base  & On vs Off       & Microwave               &  2 classes,  2 videos   \\
\bottomrule
\end{tabular}
    \begin{tablenotes}
      \item[a] 6 shared with each of \hyperref[tab:action-problems-real-world]{Apple $\times$ \{Counter Top, Freezer, Microwave, Sink\}}
      \item[b] 6 shared with each of \hyperref[tab:action-problems-real-world]{Butter Knife $\times$ \{Counter Top, Freezer, Microwave, Sink\}}
      \item[c] 6 shared with each of \hyperref[tab:action-problems-real-world]{Can $\times$ \{Counter Top, Freezer, Microwave, Sink\}}
      \item[d] 6 shared with each of \hyperref[tab:action-problems-real-world]{Hammer $\times$ \{Counter Top, Freezer, Microwave, Sink\}}
      \item[e] 6 shared with each of \hyperref[tab:action-problems-real-world]{Mug $\times$ \{Counter Top, Freezer, Microwave, Sink\}}
      \item[f] 4 shared with Pick from Counter Top, 4 shared with Pick from Table
      \item[g] 8 shared with Pick from Counter Top, 8 shared with Pick from Table
    \end{tablenotes}
  \end{threeparttable}
\vspace{0.3cm}
\caption{Number of classes and videos per each level 1 problem set testing object and object property understanding in the real world environment.}
\label{tab:object-problems-real-world}
\end{table}

\begin{table}[ht]
\centering
\begin{tabular}{llll}
\toprule
Level      & Group       & Task                            & Classes, videos per problem \\
\midrule
Level 2    & Permutation &  (8 problems)                   & 2 classes, 2 videos   \\
Levels 3, 4, 5, 8  & Permutation &  (8 problems per level)         & 3 classes, 3 videos   \\
Levels 2, 3, 4, 5, 8  & Remix       &  (3 problems per level)         & 9 classes, 9 videos   \\
\bottomrule
\end{tabular}
\vspace{0.3cm}
\caption{Number of classes and videos per each multiple-action problem set in the real world environment.}
\label{tab:list-high-level-problems-real-world}
\end{table}

\begin{table}[ht]
\centering
\begin{tabular}{lll}
\toprule
Group       & Task                              & Classes, videos per problem \\
\midrule
Real world comparison &  Pick Object from Table & 3 classes, 9 videos         \\
\bottomrule
\end{tabular}
\vspace{0.3cm}
\caption{Virtual home videos that closely match real videos.}
\label{tab:virtual-home-direct-comparison}
\end{table}

\begin{table}[ht]
\centering
\begin{tabular}{lll}
\toprule
Group              & Task                     & Classes, videos per problem \\
\midrule
Chaotic environment & Opening or closing      & 2 classes, 200 videos       \\
Virtual home comparison & Pick object from table & 3 classes, 9 videos      \\
Object interaction & Pinned or not            & 2 classes,  18 videos       \\
Object tracking    & Object identity          & 2 classes, 12 videos        \\
Object tracking    & Relative position        & 3 classes,  9 videos        \\
Object tracking    & Scramble 2 avocados      & 2 classes,  4 videos        \\
Object tracking    & Scramble 2 chopsticks    & 2 classes,  4 videos        \\
Object tracking    & Scramble 2 potatoes      & 2 classes,  4 videos        \\
Object tracking    & Scramble 2 tomatoes      & 2 classes, 10 videos        \\
Object tracking    & Scramble 3 avocados      & 2 classes,  4 videos        \\
Object tracking    & Scramble 3 chopsticks    & 2 classes,  4 videos        \\
Object tracking    & Scramble 3 potatoes      & 2 classes,  4 videos        \\
Object tracking    & Scramble 3 tomatoes      & 2 classes,  5 videos        \\
Object tracking    & Scramble 4 avocados      & 2 classes,  4 videos        \\
Object tracking    & Scramble 4 chopsticks    & 2 classes,  4 videos        \\
Object tracking    & Scramble 4 potatoes      & 2 classes,  4 videos        \\
Object tracking    & Scramble 4 tomatoes      & 2 classes,  4 videos        \\
Object tracking    & Scramble 5 chopsticks    & 2 classes,  4 videos        \\
Object tracking    & Scramble 5 tomatoes      & 2 classes, 15 videos        \\
\bottomrule
\end{tabular}
\vspace{0.3cm}
\caption{Number of classes and videos in level 1 problem sets from the real world environment that do not replicate tasks from the virtual home.}
\label{tab:list-extrapyramidal-problems-real-world}
\end{table}

\section{Experimental setup details} \label{app:experimental-setup-details}

We evaluate GPT-4o in a few-shot manner, using the following interaction scheme:

\begin{enumerate}
    \item First conversation, to get frame-by-frame descriptions: \begin{enumerate}
        \item First and only message: system prompt (figure \ref{fig:system-prompt}) + frame-by-frame prompt (figure \ref{fig:frame-by-frame-prompt})
    \end{enumerate}
    \item In a second conversation, separate from the first one: \begin{enumerate}
        \item First message: The same system prompt (figure \ref{fig:system-prompt}) + class match prompt (figure \ref{fig:class-match-prompt})
        \item Second message, same conversation, after GPT-4o replies: scoring prompt (figure \ref{fig:scoring-prompt})
    \end{enumerate}
\end{enumerate}

We found this prompting setup to yield significantly better results than forcing the model to do everything (describing the frames, then matching the descriptions to classes) in a single forward pass. We extract the scores from the last response which has predictable easy-to-parse format. For the full prompts see \cref{fig:system-prompt,fig:scoring-prompt,fig:class-match-prompt,fig:frame-by-frame-prompt}.

Figures \ref{fig:minecraft-frame-by-frame-prompt}, \ref{fig:minecraft-multilabel-scoring-prompt} and \ref{fig:minecraft-multiclass-scoring-prompt} mention the prompts used within Minecraft environment.

\lstset{
    basicstyle=\small\ttfamily,
    frame=tb,
    basewidth=4.5pt, 
    breaklines,
    linewidth=\textwidth
}

\begin{figure}[ht]
\centering
\begin{lstlisting}
You are an autoregressive language model that has been fine-tuned with instruction-tuning and RLHF. You carefully provide accurate, factual, thoughtful, nuanced answers, and are brilliant at reasoning. Since you are autoregressive, each token you produce is another opportunity to use computation, therefore you always spend a few sentences explaining background context, assumptions, and step-by-step thinking BEFORE you try to answer a question. You always use precise, plain scientific language, without unneeded flourish. You provide details where it might help the explanation.
\end{lstlisting}
\caption{The system prompt, used in both requests.}
\label{fig:system-prompt}
\end{figure}

\begin{figure}[ht]
\centering
\begin{lstlisting}
You will be given {{n_frames}} frames from a first-person video. The frames are given to you in chronological order.

{{task.prompt_gpt}}

{{task.example_gpt}}

Input frames. Describe each frame separately.

{{frames}}
\end{lstlisting}
\caption{The prompt used to obtain frame-by-frame descriptions. \emph{n\_frames} depends on the model setting (by default, 16), \emph{task.prompt\_gpt} and \emph{task.example\_gpt} are problem-specific, the latter only being used in 1-shot mode (which is the default). \emph{frames} refer to the actual frames, which are base64 encoded and passed through the OpenAI API as detailed in their documentation.}
\label{fig:frame-by-frame-prompt}
\end{figure}

\begin{figure}[ht]
\centering
\begin{lstlisting}
Consider the following sequence of frame-by-frame descriptions:

```
{{frame_descriptions}}
```

Break down each of the following summaries into individual steps, and mention what frames or frame ranges each step matches in parentheses after each step. Note even partial matches, e.g. matching a kind of action (put, pick, ...) even though the object might be incorrect. Also provide a one-sentence commentary on how well each summary matches the sequence of the frame descriptions. In the commentary, the most important thing is to match the kinds of actions performed and their order. For example if put (of anything) was described before a pick (of anything) in the frames, maintaining this order in the summary is more important than getting the exact object right. Do not comment on the overall quality of the summaries.

Summaries, given in the format `- (label) summary`:
{{class_list}}
\end{lstlisting}
\caption{The prompt used to make GPT-4o think step by step about how well each class description matches the frame descriptions it produced in the first request. The \emph{frame\_descriptions} are the generated frame-by-frame descriptions, \emph{class\_list} is the list of label-description pairs from the task yaml definition. The raw frames are not supplied to the model here anymore.}
\label{fig:class-match-prompt}
\end{figure}

\begin{figure}[ht]
\centering
\begin{lstlisting}
 Based on your findings in the previous section, score the summaries from 0 to 5, where 0 is "likely does not describe the video" and 5 is "among the given options, this one most likely describes the video". Make sure to score each summary individually. At least one score should be non-zero, even if it's not a perfect match. Whatever scores you pick, there !must be! exactly one summary with the highest score. Follow the format below, verbatim:

```
- (label) score
```
\end{lstlisting}
\caption{The prompt used to obtain the final scores in a predictable format.}
\label{fig:scoring-prompt}
\end{figure}

\begin{figure}[ht]
\centering
\begin{lstlisting}
  # TASK
  You will be given several frames from a Minecraft footage. The frames are given in chronological order.
  First, describe each frame. Focus on what activities the player is performing in each frame and what the player interacts with. Be brief and to the point.
  # EXAMPLE OF FRAME DESCRIPTIONS
  Input: [ten frames]

  Assistant:
  1. We see a flat grassy area with small houses and trees nearby. Player holds an oak fence block in their hands.
  2. Same scene from a sligthly different angle.
  3. Player breaks a tuft of grass using an oak fence block.
  4. Player proceeds removing grass.
  5. Player places a fence block next to a house.
  6. Player places more fence blocks, building a straight fence line.
  7. Player turns around and finds a pig.
  8. Player places one more fence block in the line.
  9. Player breaks another tuft of grass.
  10. Player proceeds removing grass.
\end{lstlisting}
\caption{The prompt used to obtain frame-by-frame descriptions for Minecraft videos. \emph{frames} refer to the actual frames, which are base64
encoded and passed through the OpenAI API as detailed in their documentation.}
\label{fig:minecraft-frame-by-frame-prompt}
\end{figure}

\begin{figure}[ht]
\centering
\begin{lstlisting}
Then, given the original frames and your description, score the following potential video descriptions from 0 to 1 based on how well they describe the video you've seen. Feel free to use values between 0 and 1, too. There could be more than one 'correct' description with score 1. The descriptions are given in the following format:

- (label) description

Options:

{classes}

The final scores should be in the following format, verbatim:

```
- (label) your score
```

Be sure not to alter the label in any way, since we will use it to match your scores to the potential descriptions we've given you.
\end{lstlisting}
\caption{The prompt used to obtain the final scores in a predictable format for Minecraft videos in \textbf{multilabel} setting.}
\label{fig:minecraft-multilabel-scoring-prompt}
\end{figure}

\begin{figure}[ht]
\centering
\begin{lstlisting}
Then, given the original frames and your description, score the following potential video descriptions from 0 to 1 based on how well they describe the video you've seen. Feel free to use values between 0 and 1, too. There should be exactly one 'correct' description with score 1. The descriptions are given in the following format:

- (label) description

Options:

{classes}

The final scores should be in the following format, verbatim:

```
- (label) your score
```

Be sure not to alter the label in any way, since we will use it to match your scores to the potential descriptions we've given you.
\end{lstlisting}
\caption{The prompt used to obtain the final scores in a predictable format for Minecraft videos in \textbf{multiclass} setting.}
\label{fig:minecraft-multiclass-scoring-prompt}
\end{figure}

\section{Full results} \label{app:full-results}

\begin{figure}[ht] 
    \centering
    \begin{subfigure}{\textwidth}
        \includegraphics[width=\textwidth]{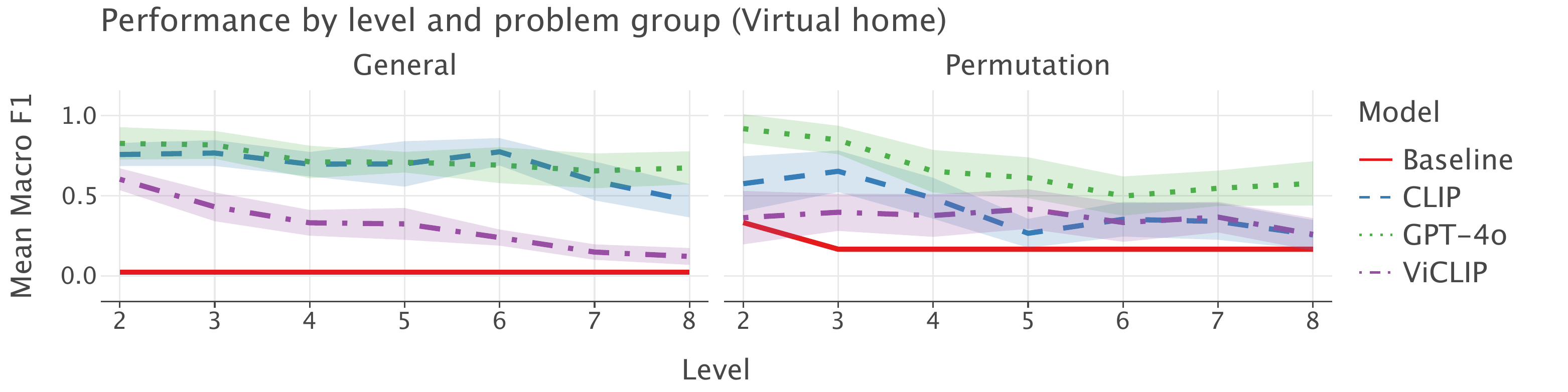}
        \caption{
            \textbf{Longer tasks see a large drop in performance.}
        }
    \end{subfigure}
    \vspace{0.5cm} 
    
    \begin{subfigure}{\textwidth}
        \includegraphics[width=\textwidth]{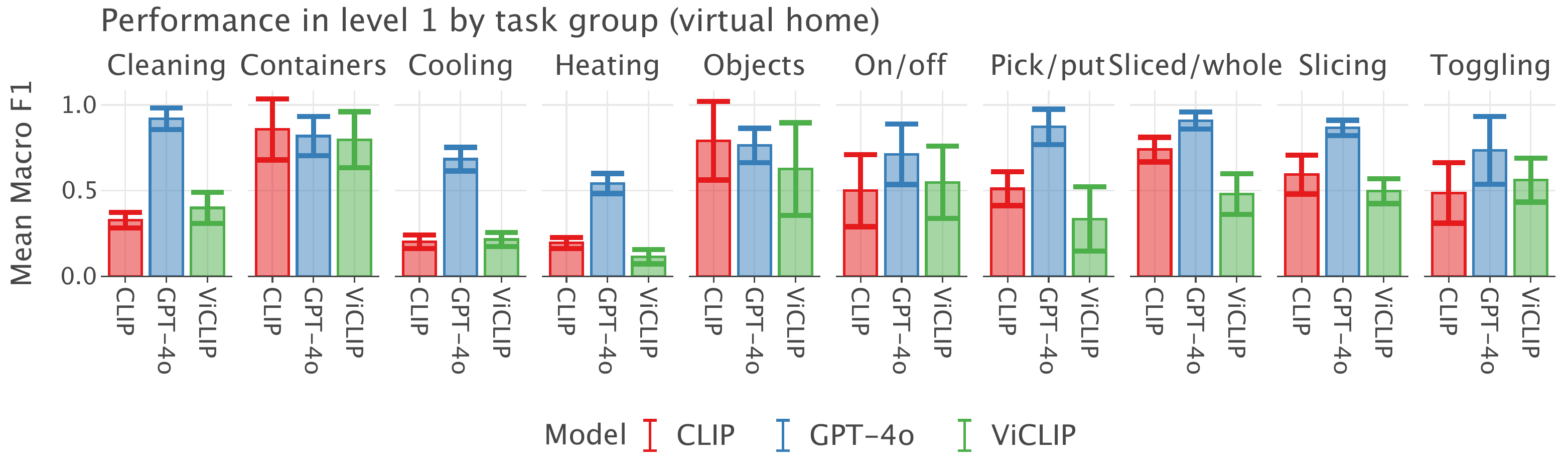}
        \caption{
            \textbf{Action understanding is difficult.}
            While the performance on object recognition tasks is high for all models, only {\gptfour}o has acceptable performance on action understanding.
            Even there, the more complex actions like cooling and heating lead to large performance drops.
        }
    \end{subfigure}
    \caption{Performance overview in the virtual home environment. Ranges are 95\% C.I.}
    \label{fig:results-full-virtual-home}
\end{figure}

\begin{figure}[ht]
    \centering
    \begin{subfigure}{\textwidth}
        \includegraphics[width=\textwidth]{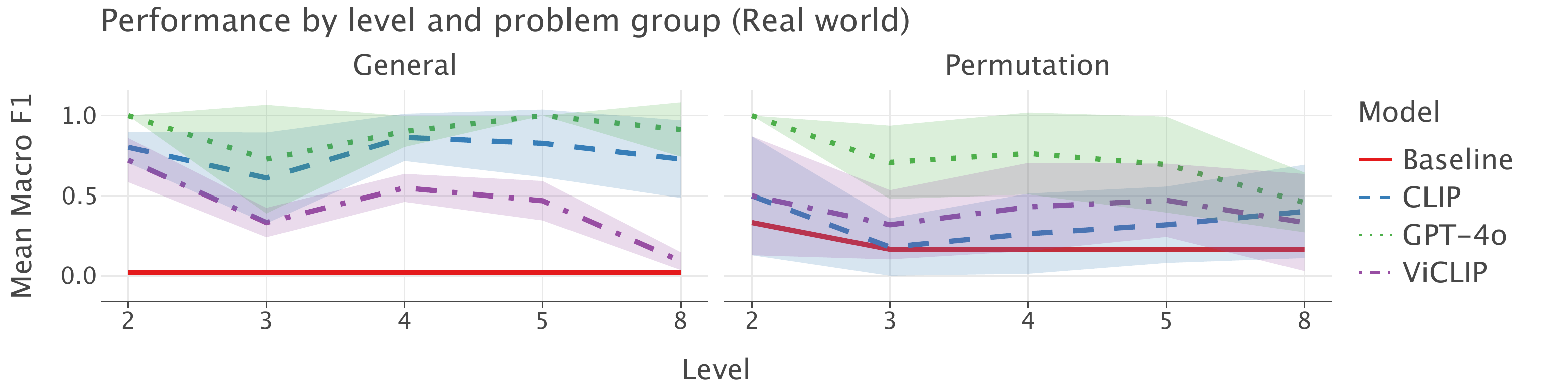}
        \caption{
            \textbf{Longer tasks do not see a large drop in performance in the real world environment.} 
            The high variance makes it difficult to draw any conclusions.
        }
    \end{subfigure}
    \vspace{0.5cm} 
    
    \begin{subfigure}{0.7\textwidth}
        \includegraphics[width=\textwidth]{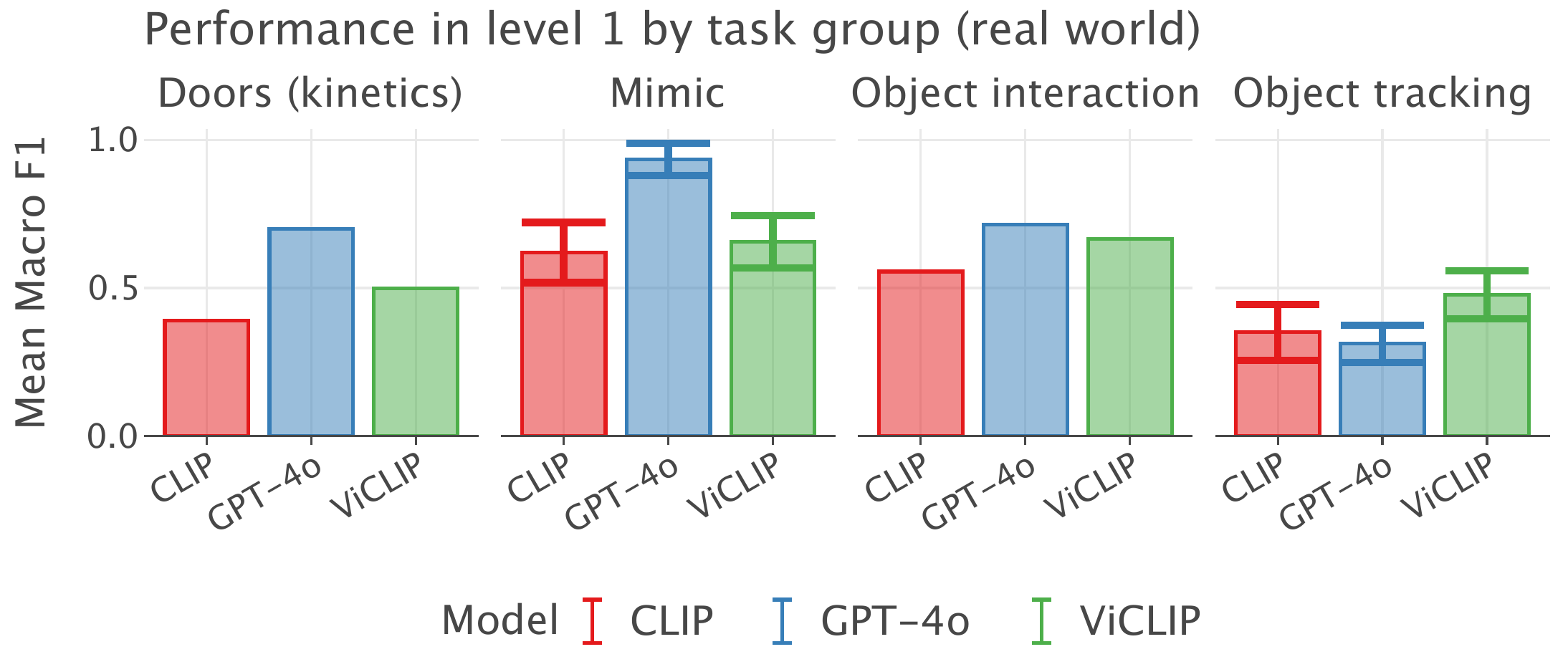}
        \caption{
            \textbf{Object tracking is very difficult.}
        }
    \end{subfigure}
    \caption{Performance overview in the real world environment. Ranges are 95\% C.I.}
    \label{fig:results-full-real-world}
\end{figure}

\begin{figure}[ht] 
    \centering
    \begin{subfigure}[t]{0.5\textwidth}
        \includegraphics[width=\textwidth]{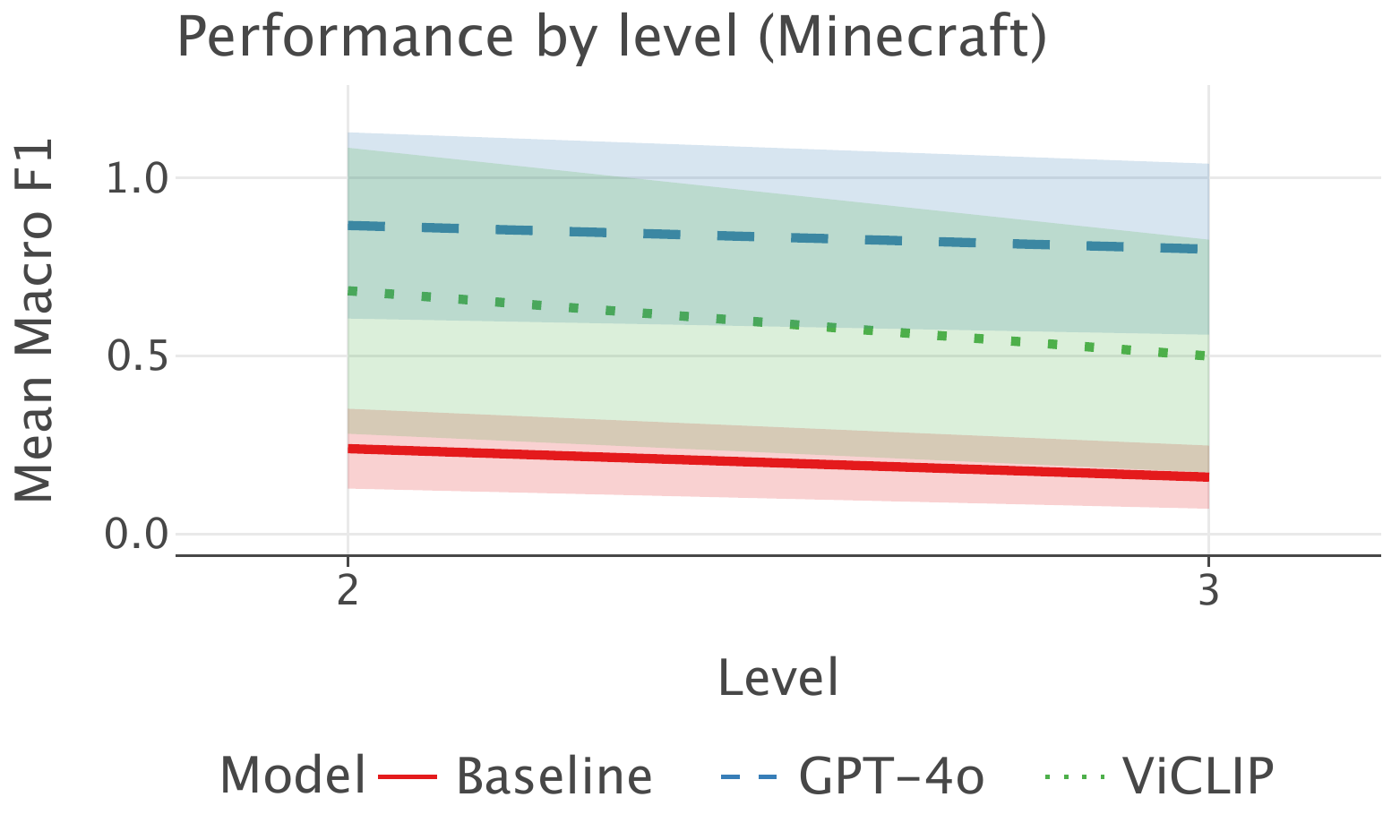}
        \caption{
            In Minecraft, we only tested a handful of multiple-action problem sets. It is difficult to draw any conclusions.
        }
    \end{subfigure}
    \hspace{0.05\textwidth}
    \begin{subfigure}[t]{0.39\textwidth}
        \centering
        \includegraphics[width=0.45\textwidth]{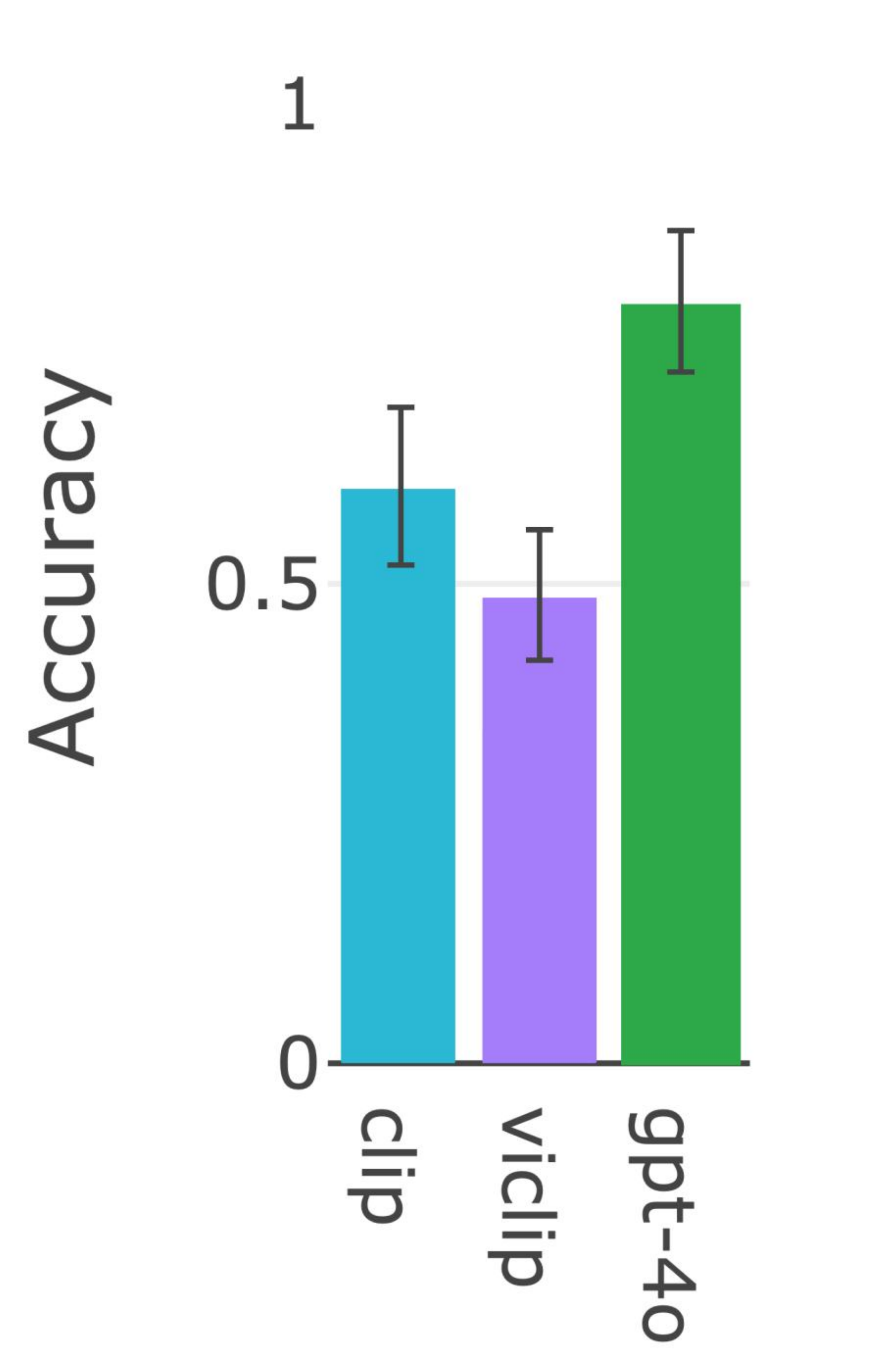}
        \caption{
           Even in Minecraft's level 1, {\gptfour}o is clearly better than the other two models.
        }
    \end{subfigure}
    \caption{Performance overview in the Minecraft environment. Ranges are 95\% C.I.}
    \label{fig:results-full-minecraft}
\end{figure}

\begin{figure}    
    \includegraphics[width=\textwidth]{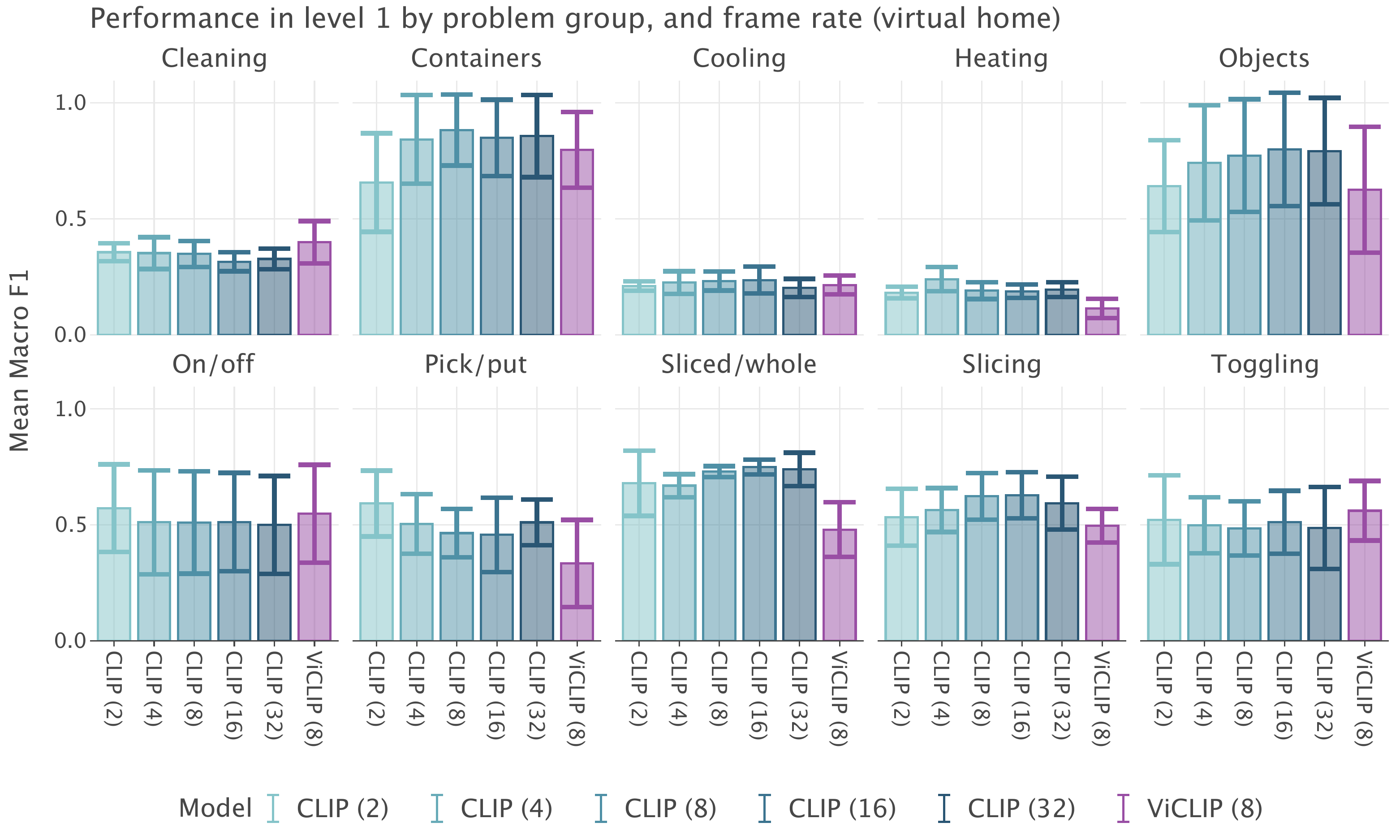}

    \caption{
        \textbf{Frame rate does not play a large role in level 1 tasks.}
        The performance of {\clip} and {\viclip} on different groups of problem sets in the virtual home environment, after changing the number of {\clip} input frames.
        Ranges are 95\% C.I.
    }
    \label{fig:results-full-clips}
\end{figure}

Refer to \cref{fig:results-full-virtual-home} for expanded results in virtual home, \cref{fig:results-full-real-world} for results in the real world environment, \cref{fig:results-full-minecraft} for Minecraft, and \cref{fig:results-full-clips} for complete results comparing {\clip} models with different frame rates.

\begin{figure}[ht]
    \centering
\begin{subfigure}{0.4\textwidth}
\includegraphics[width=\textwidth]{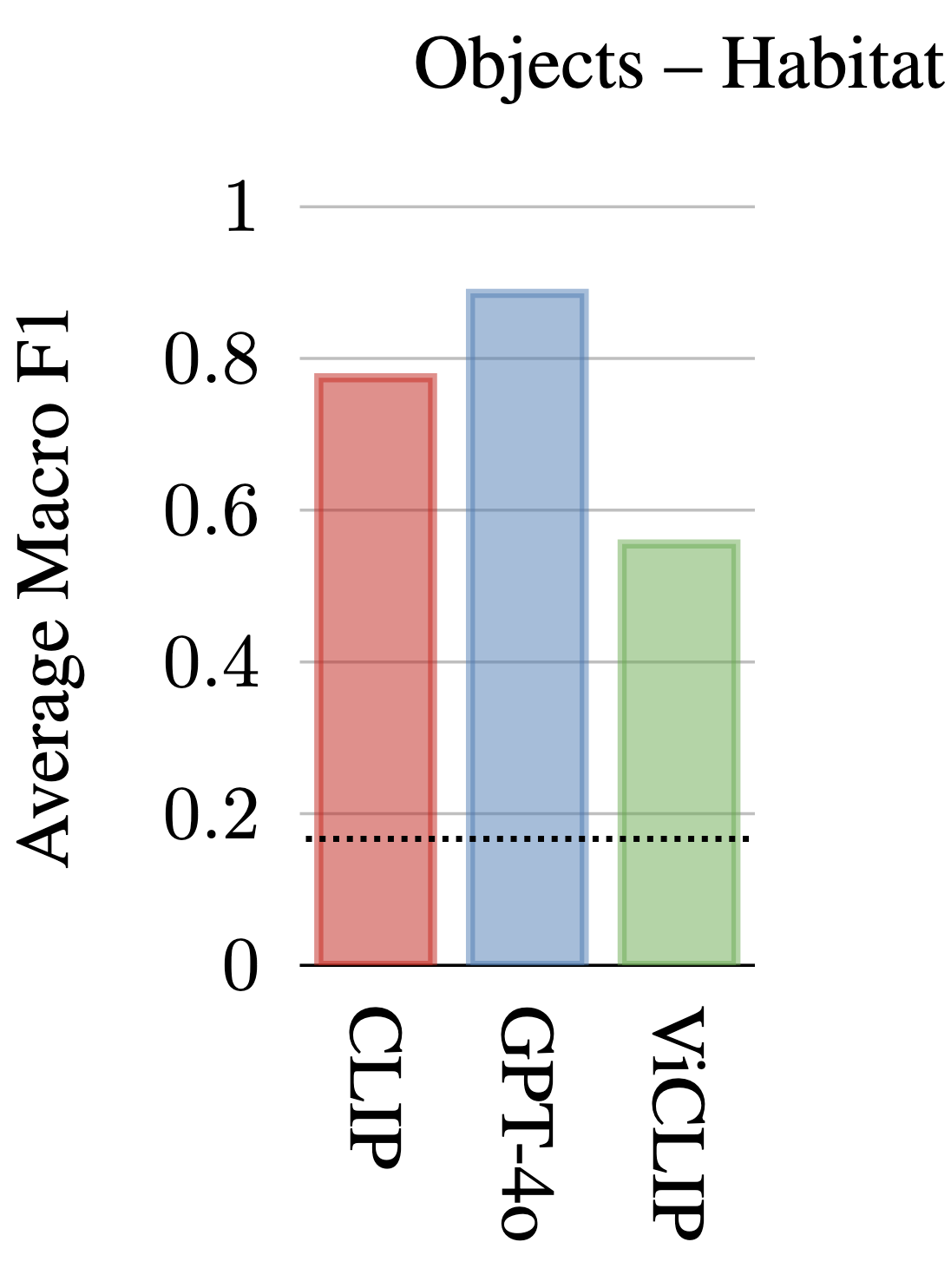}
\end{subfigure}
\hspace{1cm}
\begin{subfigure}{0.4\textwidth}
\includegraphics[width=\textwidth]{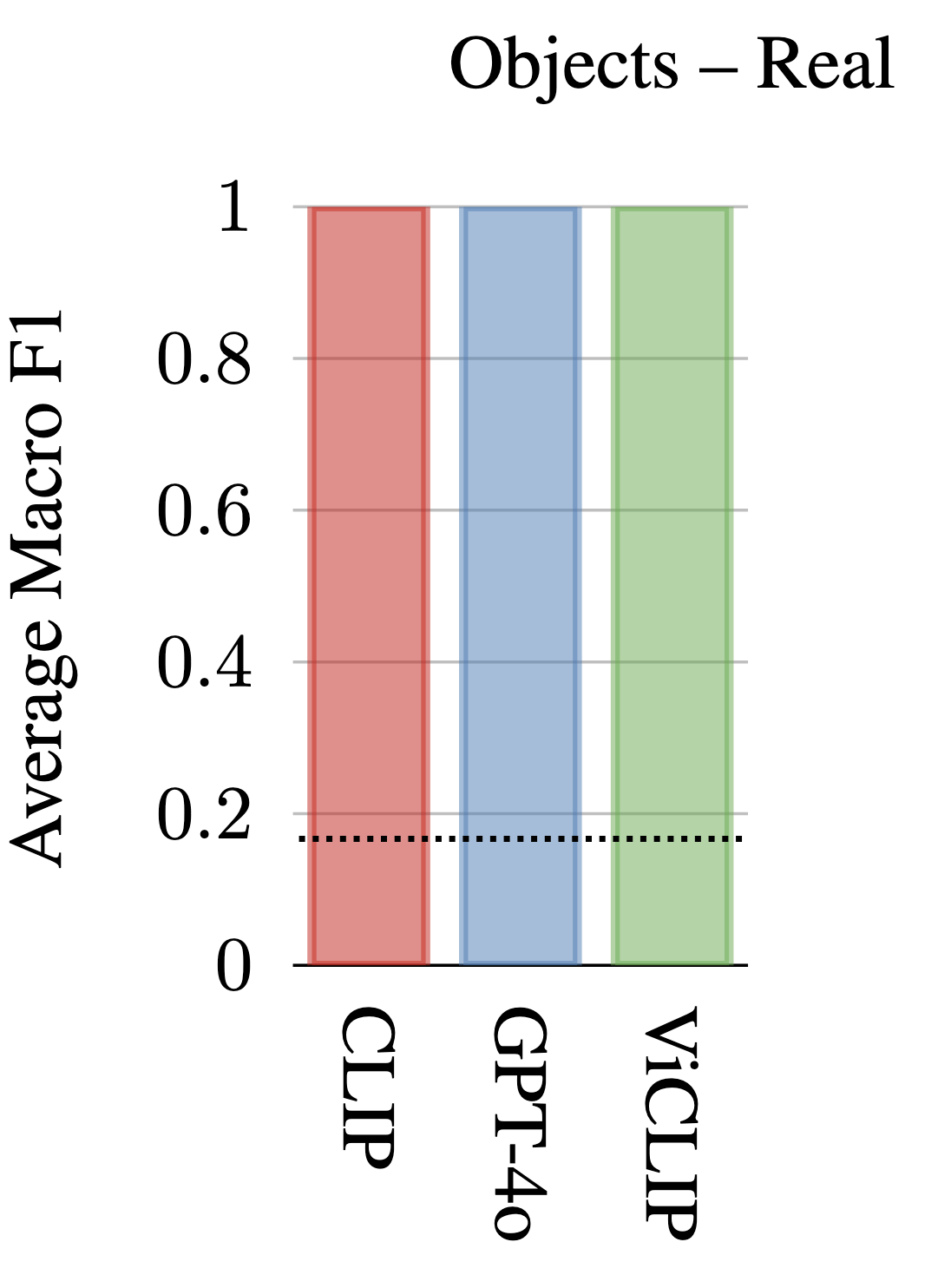}
\end{subfigure}
\caption{Performance on identifying which object was picked up from a cluttered table in the virtual home environment Habitat~\cite{puigHabitatCoHabitatHumans2023a} and real-world videos which are close recreations of the Habitat ones, with similar movements, angles, and background objects.
None of the models made mistakes on any of the 9 real videos, but all made mistakes on at least one virtual home video.}
\label{fig:objects-habitat-vs-real}
\end{figure}

\end{document}